\documentclass[runningheads]{llncs}

\usepackage{eccv}

\usepackage{graphicx}
\usepackage{booktabs}

\usepackage{tabularx}
\usepackage{array}
\usepackage{amsmath}
\usepackage{makecell}
\usepackage[accsupp]{axessibility}

\usepackage{hyperref}

\usepackage{orcidlink}

\usepackage{multirow}                

\usepackage{pifont}

\usepackage{threeparttable}

\begin{document}

\title{FFAvatar: Feed-Forward 4D Head Avatar Reconstruction from Sparse Portrait Images}

\titlerunning{FFAvatar}


\author{
Jianjiang Yao\orcidlink{0009-0003-0592-6945}
\and
Ke Xian\thanks{Corresponding author.}\orcidlink{0000-0002-0884-5126}
\and
Renxiang Dai\orcidlink{0000-0002-5695-1986}
\and
Robert Caiming Qiu\orcidlink{0000-0002-1154-1836}
}

\authorrunning{J.~Yao et al.}


\institute{
School of Electronic Information and Communications,\\
Huazhong University of Science and Technology, Wuhan 430074, China\\
\email{\{jjyao, kxian, rxdai, caiming\}@hust.edu.cn}
}

\maketitle
\begin{center}
\includegraphics[width=\textwidth]{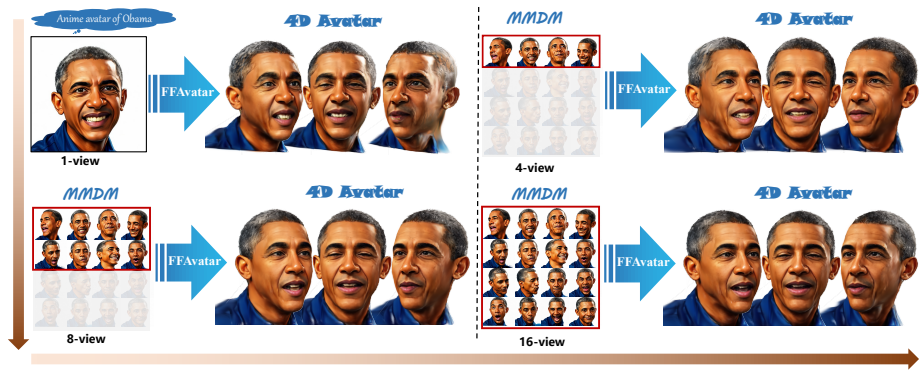}

\captionof{figure}{\textbf{FFAvatar.} 
We present FFAvatar, a feed-forward framework that reconstructs high-fidelity 4D portrait avatars from one or more reference portrait images.
The reference images can be virtual characters synthesized by large-scale generative models from text prompts. 
Furthermore, by leveraging diffusion-based MMDM~\cite{taubner2025cap4d} to synthesize novel-view and novel-expression 2D images, our approach further enhances reconstruction quality and view-consistent appearance.}
\label{fig:teaser}
\end{center}

\begin{abstract}
We present FFAvatar, a Transformer-based 3D Gaussian framework for fast construction of high-quality and animatable 4D head avatars from one or more reference portrait images. Unlike existing feed-forward approaches that require a fixed number of input views, FFAvatar supports \emph{incremental reconstruction}, progressively refining the avatar representation as additional reference images become available.
At the core of our method is an alternating attention mechanism that disentangles identity appearance from expression and viewpoint variations, enabling the reconstruction of a canonical 3D appearance that remains consistent across poses and facial expressions. To balance visual fidelity and computational efficiency, we introduce a sparse-to-dense learning paradigm. Coarse appearance features are first learned using sparse primitives anchored to the FLAME vertex level and are subsequently densified in the UV domain to capture fine-grained geometric and texture details.
We further propose a plug-and-play motion refinement module that enables subject-specific dynamic personalization by modeling residual motion beyond parametric deformation. Extensive experiments demonstrate that FFAvatar efficiently produces high-fidelity and controllable 4D head avatars, achieving superior flexibility, driving efficiency, and identity-consistent rendering across diverse expressions and viewpoints. \textbf{Project Page:} \href{https://jj-yao.github.io/ffavatar/}{https://jj-yao.github.io/ffavatar/}
\end{abstract}

\section{Introduction}
\label{sec:intro}

High-quality 4D avatar head reconstruction plays a critical role in a wide range of applications, including virtual reality, digital humans, telepresence, and immersive content creation. An ideal 4D avatar head should faithfully preserve a person’s identity while supporting realistic and temporally stable facial dynamics across diverse expressions, head poses, and viewpoints.

Despite the remarkable progress in neural rendering~\cite{nerf,instant} and avatar modeling~\cite{athar2022rignerf,gafni2021dynamic,hong2022headnerf,kania2022conerf,park2021nerfies}, existing approaches still suffer from several key limitations.
\ding{182} \textbf{Limited generalization under sparse observations.}
Many existing methods rely on densely captured multi-view observations and perform object-specific optimization~\cite{xiang2024flashavatar,qian2024gaussianavatars,guo2021ad,li2023efficient,aneja2024gaussianspeech,cho2024gaussiantalker,li2024talkinggaussian,xu2025vasa,xu2024gaussian}. While these approaches can achieve high-quality reconstruction, they generalize poorly under single- or few-shot settings.
Feed-forward 3D reconstruction provides a promising alternative for sparse-input scenarios. 
However, most existing feed-forward 4D avatar methods~\cite{deng2024portrait4dv2,he2025lam,chu2024generalizable,ye2024real3d,deng2024portrait4d} mainly focus on the single-reference setting, where limited viewpoint cues often lead to incomplete geometry and ambiguous appearance.
Some recent methods~\cite{kirschstein2025avat3r,wu2025fastavatar,ji2026fastgha} extend feed-forward reconstruction to few-shot settings by constructing multiple view-dependent canonical subspaces from the input images.
However, such a multi-canonical design tends to introduce redundant Gaussian representations and higher computational overhead as more reference views are used, limiting its scalability in sparse-view reconstruction.
\ding{183} \textbf{Entanglement between identity appearance and motion.}
Effectively disentangling identity-related appearance from facial expressions and head pose remains a challenging problem.
In many existing approaches, identity features are intertwined with motion-related variations, which often leads to visual artifacts, identity drift, and rendering inconsistencies when synthesizing novel expressions or viewpoints~\cite{he2025lam,chu2024generalizable}.
\ding{184} \textbf{Trade-off between representation fidelity and efficiency.}
High-fidelity avatar rendering often relies on dense UV representations or a large number of 3D Gaussians~\cite{xiang2024flashavatar,jiang2025uv,kirschstein2024gghead,he2025lam,chu2024generalizable}.
Although these dense representations significantly improve rendering quality, they also introduce substantial memory consumption and computational overhead.

To address these challenges, we propose FFAvatar, a novel feed-forward framework for incremental 4D portrait avatar reconstruction from one or more reference portrait images.
Our method is built upon three key strategies:
\ding{182} \textbf{Flexible multi-view feature aggregation and unified canonical field construction.}
To address these limitations, we introduce an alternating attention mechanism, inspired by VGGT~\cite{wang2025vggt}, to aggregate identity-aware appearance information from a variable number of reference images while disentangling expression and viewpoint variations. Instead of maintaining multiple view-dependent canonical subspaces, we leverage the FLAME prior~\cite{li2017learning} to construct a unified global canonical Gaussian field, where the aggregated global appearance representation is injected through cross-modal alignment.
This unified design avoids redundant view-dependent representations and enables high-quality 4D head avatar reconstruction in a single forward pass, leading to improved flexibility, efficiency, and scalability under sparse observations.
\ding{183} \textbf{Decoupled modeling of identity and motion.}
We explicitly model identity appearance and motion variations separately.
For identity modeling, the global identity representation produced by the alternating attention module is decoded into an expression- and view-independent 3D appearance.
For motion modeling, we introduce a plug-and-play motion refinement module that captures dynamic motion patterns beyond the expressiveness of the FLAME template, enabling subject-specific dynamic personalization.
\ding{184} \textbf{Sparse-to-dense hierarchical representation learning.}
To balance representation fidelity and computational efficiency, we propose a sparse-to-dense learning paradigm.
The model first learns coarse appearance features using sparse primitives anchored to the FLAME vertex level, and then progressively densifies the representation in the UV domain to capture high-frequency geometric and texture details.
Compared with directly optimizing uniformly dense Gaussian representations, this hierarchical design significantly reduces computational cost while maintaining high rendering quality.

As illustrated in Fig.~\ref{fig:teaser}, our framework reconstructs high-fidelity 4D avatars from one or more reference portrait images, including AI-generated virtual portraits. Unlike existing feed-forward avatar head reconstruction methods~\cite{kirschstein2025avat3r,deng2024portrait4d,deng2024portrait4dv2,he2025lam,chu2024generalizable,ye2024real3d}, \textbf{FFAvatar supports incremental reconstruction}, allowing the avatar representation to be progressively refined as additional images become available. 
This property enables broader applicability in practical scenarios. 
For instance, by leveraging the generative capability of diffusion models~\cite{wu2025cat4d,taubner2025cap4d}, our method can synthesize high-quality 4D avatars even under a single-image setting, which is difficult to achieve with previous feed-forward single-image reconstruction approaches~\cite{deng2024portrait4d,deng2024portrait4dv2,he2025lam,chu2024generalizable,ye2024real3d}.

Overall, we make the following contributions:
\begin{itemize}
    \item We present {FFAvatar}, an incremental feed-forward framework that enables rapid reconstruction of high-fidelity 4D head avatars from one or more reference portrait images.

    \item We propose a {sparse-to-dense hierarchical feature learning paradigm}, learning coarse yet semantically stable features at the sparse FLAME vertex level and densifying them in UV space to capture high-frequency geometric and texture details, achieving a better trade-off between efficiency and fidelity.

    \item We introduce a motion refinement module that enables subject-specific dynamic personalization. Built upon parametric deformation, this module further enhances motion realism by refining identity-dependent dynamic details.

\end{itemize}

\section{Related Work}
\subsection{Mesh-based Facial Reenactment}
Mesh-based facial reenactment has been widely studied~\cite{doukas2021headgan,drobyshev2022megaportraits,yang2022face2face,khakhulin2022realistic,tewari2020stylerig,yao2020mesh,zeng2022fnevr,feng2021learning,danvevcek2022emoca,zielonka2022towards,filntisis2022visual,chu2025artalk} for explicit face animation using parametric 3D models. Existing methods can be broadly categorized into deformation-based and graphics-based approaches.
Deformation-based methods warp source images or meshes by estimating explicit motion fields~\cite{doukas2021headgan,yang2022face2face,yao2020mesh}, based on mesh correspondences, geometry-guided dense flows, or sparse 3D landmark motions.
However, they often degrade under large head rotations or significant geometric changes, where explicit deformation modeling becomes less robust.
Graphics-based approaches~\cite{feng2021learning,danvevcek2022emoca,zielonka2022towards,filntisis2022visual} instead reconstruct animatable head meshes and synthesize images through differentiable or hybrid rendering pipelines.
Early works rely on parametric 3D morphable models to estimate geometry and appearance from monocular inputs, while recent learning-based methods recover detailed geometry, expression-dependent deformations, and identity-specific shapes from in-the-wild data.
However, despite improved fidelity and expression realism, mesh-based pipelines~\cite{tewari2020stylerig,yao2020mesh,feng2021learning,danvevcek2022emoca,zielonka2022towards,filntisis2022visual,huang2023simhmr} still depend on costly differentiable or hybrid rendering systems and often struggle to capture fine-grained facial details.

\subsection{Feed-Forward 3D Reconstruction Models}

Feed-forward 3D reconstruction methods aim to recover 3D geometry and appearance directly through a single forward pass of a neural network, avoiding expensive per-instance optimization at inference time~\cite{hong2023lrm,ren2024l4gm,liang2024feed,yang2024storm,qi2025predicting,wang2025vggt,shen2025fastvggt}. Early works primarily focused on reconstructing coarse 3D shapes from single images using volumetric~\cite{bai2023learning,giebenhain2024mononphm,zielonka2023instant}, point-cloud~\cite{zheng2023pointavatar}, or mesh-based~\cite{athar2024bridging,grassal2022neural,ichim2015dynamic,zheng2022avatar,ma2021pixel} representations, demonstrating the feasibility of fast inference but often suffering from limited geometric detail and poor generalization to complex poses or expressions.
With the advent of neural implicit representations~\cite{nerf,instant}, feed-forward models have been extended to learn continuous geometry and appearance fields from images~\cite{hong2023lrm}. Approaches based on signed distance functions or neural radiance fields have significantly improved reconstruction quality, yet many of them~\cite{xiang2024flashavatar,zheng2023pointavatar,qian2024gaussianavatars,ma20243d,chen2024monogaussianavatar,wang2025gaussianhead} still require per-subject fine-tuning or rely on multi-view inputs~\cite{kirschstein2025avat3r} during inference, limiting their practical efficiency. More recently, feed-forward neural rendering frameworks have explored hybrid representations that combine explicit geometry with learned appearance features~\cite{wang2025vggt,shen2025fastvggt}, enabling faster inference while maintaining high visual fidelity.

\subsection{3D Avatar Head Reconstruction}

Reconstructing high-quality 3D avatar heads with realistic appearance and expressive dynamics has long been a fundamental problem in computer vision and graphics~\cite{taubner2025cap4d}.
Recent advances in 3D Gaussian Splatting~\cite{kerbl20233d} have significantly improved the realism of head reconstruction. By representing scenes with 3D Gaussian ellipsoids, these methods can effectively model fine geometric details and view-dependent appearance. However, most existing approaches~\cite{chen2024monogaussianavatar,xiang2024flashavatar,zheng2023pointavatar,peng2024synctalk,aneja2024gaussianspeech,li2024talkinggaussian} require per-subject optimization and dense multi-view data acquisition, resulting in high computational cost and limited scalability. In addition, identity appearance is often tightly coupled with facial expressions and head poses, which can lead to artifacts or identity drift when synthesizing novel expressions or viewpoints.
To address these challenges, recent studies have explored hybrid representations that combine explicit geometric priors with learned appearance models. Examples include mesh-based neural textures and more recent Gaussian-based avatar representations, which enable efficient differentiable rendering and improved visual fidelity. Although these representations reduce rendering overhead, current methods still frequently rely on iterative optimization~\cite{xiang2024flashavatar,zheng2023pointavatar,qian2024gaussianavatars,ma20243d,chen2024monogaussianavatar,wang2025gaussianhead,feng2025gpavatar,li2024talkinggaussian,xu2024gaussian} or constrained input conditions~\cite{deng2024portrait4d,he2025lam,deng2024portrait4dv2,ye2024real3d,chu2024gpavatar,li2023one,li2023generalizable,ma2024cvthead,chu2024generalizable,liang2025fastavatar}, limiting their ability to perform rapid reconstruction from unconstrained data. 
Some feed-forward methods~\cite{kirschstein2025avat3r,ji2026fastgha,wu2025fastavatar} attempt to address this issue by constructing and fusing multiple view-dependent canonical subspaces from the input images. However, under multi-view input settings, their number of 3D Gaussians, creation time, and GPU memory consumption increase substantially with the number of input views.

In contrast, FFAvatar focuses on fast feed-forward 4D head avatar reconstruction by leveraging the FLAME prior to build a unified global canonical space. As a result, the number of Gaussians and the animation speed are independent of the number of input images. Our method can efficiently reconstruct a 4D head avatar in a feed-forward manner from hundreds of input images on a single A800 GPU. By explicitly disentangling identity appearance from expression and pose, and by introducing a Transformer-based global appearance aggregation mechanism within a unified canonical space, FFAvatar achieves efficient reconstruction while maintaining strong identity consistency and generalization to unseen expressions and viewpoints.

\section{Method}
\begin{figure}[h]
\centering
\includegraphics[width=1.0\linewidth]{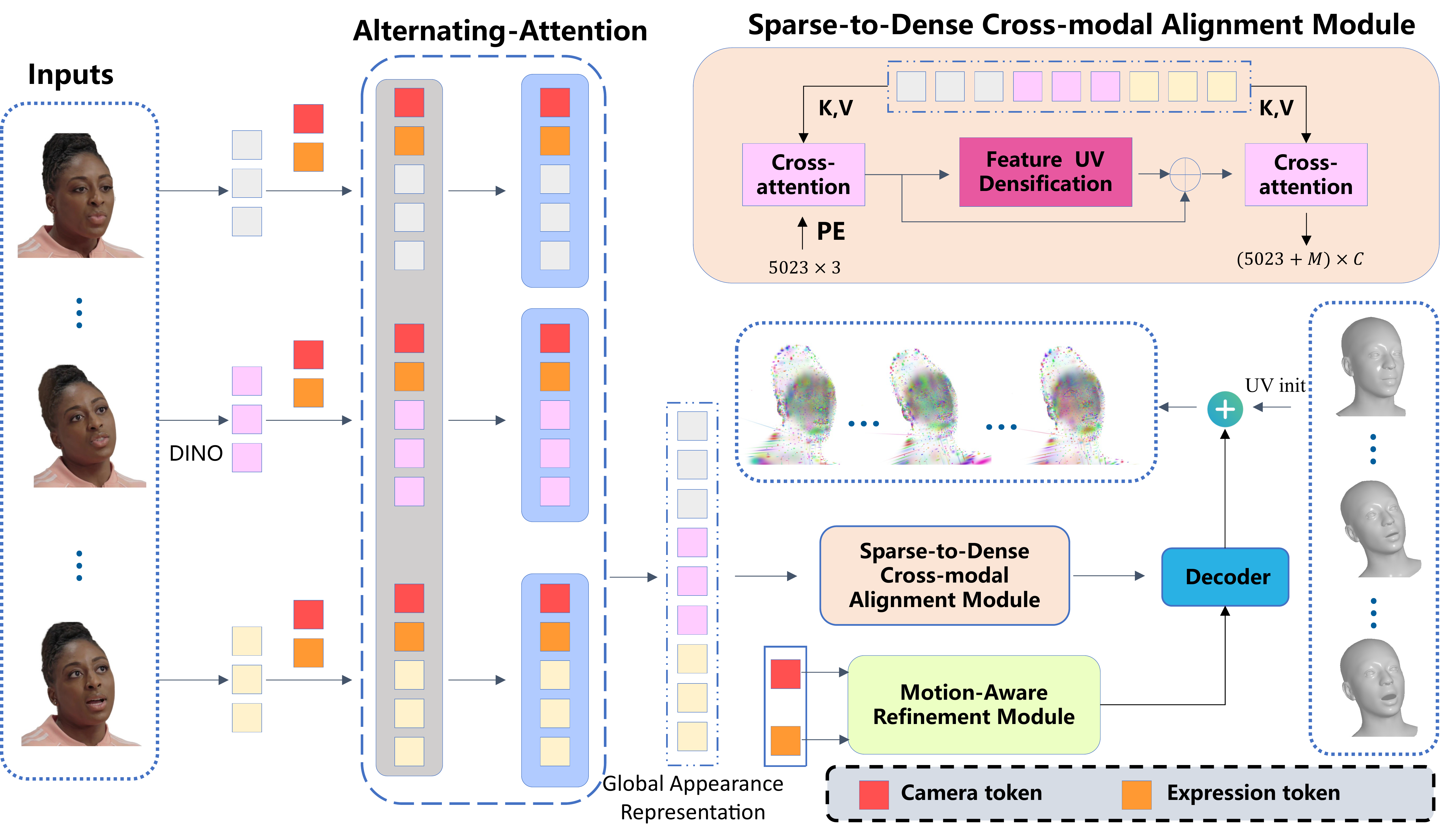}
\caption{\label{fig:ffavatar}Overview of FFAvatar. \textbf{Canonical field modeling.} Visual features extracted by DINOv3 are concatenated with camera pose and expression encodings. An alternating attention mechanism performs intra- and inter-image matching to infer a consistent global appearance representation across all inputs. This representation is then aligned with the FLAME template through the proposed Sparse-to-Dense Cross-modal Alignment Module, producing an expression- and viewpoint-invariant static 3D appearance. 
\textbf{Deformable field modeling.} Facial motions are driven by the FLAME model using standard linear blend skinning (LBS) and corrective blendshapes, and further refined by the proposed Motion-Aware Refinement Module to capture fine-grained, identity-dependent dynamic details.
}
\end{figure}

As illustrated in Fig.~\ref{fig:ffavatar}, our FFAvatar framework consists of two main stages. 
(1) \textbf{Static Appearance Canonical Field Generation}: we extract a global appearance representation from multi-view and multi-expression images, perform sparse-to-dense cross-modal alignment, and decode an expression- and viewpoint-invariant static 3D appearance. 
(2) \textbf{3D Head Avatar Animation}: the static appearance is animated using the FLAME model, followed by motion refinement through a Motion-Aware Refinement Module to capture fine-grained dynamic details.

\subsection{Static Appearance Canonical Field Generation}
\label{sec:static_canonical_field}

Given a set of head images 
$\mathcal{I} = \{ I_i \}_{i=1}^{N}$ captured under different viewpoints and expressions,
our goal is to reconstruct a viewpoint- and expression-invariant canonical 3D head appearance field
represented by 3D Gaussian primitives, denoted as $G$.

\noindent\textbf{Feature Extraction and Global Appearance Aggregation.}
For each image $I_i$, we extract dense visual features using a pretrained encoder~\cite{simeoni2025dinov3}:

\begin{equation}
I_{i}^{\mathrm{feat}} =
\operatorname{\textsc{DinoV3}}\!\left(I_i\right)
\in \mathbb{R}^{H \times W \times C}
\end{equation}
Each feature map is augmented with camera and expression embeddings, denoted as $\mathbf{Z}^{\text{cam}}$ and $\mathbf{Z}^{\text{exp}}$, respectively, and processed by an alternating attention architecture that interleaves intra-image and inter-image attention. This design aggregates identity-consistent appearance cues across images while suppressing viewpoint- and expression-specific variations. The final global appearance representation is obtained by collecting tokens from all images:
\begin{equation}
\mathbf{Z}^{\text{app}}  \gets  \operatorname{\textsc{AlterAtt}}([\{ I_{i}^{\mathrm{feat}}  \}_{i=1}^{N}, \mathbf{Z}^{\text{cam}}, \mathbf{Z}^{\text{exp}}])  \in \mathbb{R}^{N H W \times C}
\end{equation}

\noindent\textbf{Sparse-to-Dense Cross-modal Alignment.}
To balance computational efficiency and representation fidelity, we adopt a sparse-to-dense alignment strategy.  
We first initialize sparse Gaussian centers using FLAME template vertices 
$\mathbf{V}_0 \in \mathbb{R}^{5023 \times 3}$ and align them with the global appearance representation via cross-attention:
\begin{equation}
\mathbf{T}_{s}  \gets  \operatorname{\textsc{CrossAtt}}(\mathrm{PE}(\mathbf{V}_0), \mathbf{Z}^{\text{app}} )
\end{equation}

To recover high-frequency appearance details, we further densify features in the UV domain. 
Given a UV resolution $\text{uv\_size}$, the FLAME mesh is rasterized into a planar UV grid of size $\text{uv\_size} \times \text{uv\_size}$. 
Each valid UV location corresponds to a triangle on the mesh and is associated with barycentric weights defined over its three vertices. 
Dense UV features are obtained by interpolating the aligned sparse features:
\begin{equation}
\mathbf{T}_{\text{uv}}(u,v)
=
\alpha\,\mathbf{T}_s(i)
+
\beta\,\mathbf{T}_s(j)
+
\gamma\,\mathbf{T}_s(k)
\end{equation}
where $(i,j,k)$ denote the triangle vertex indices and $(\alpha,\beta,\gamma)$ are the corresponding barycentric coefficients.
Flattening all valid UV samples yields dense features
$\mathbf{T}_{\text{uv}} \in \mathbb{R}^{M \times C}$,
where $M < \text{uv\_size}^2$ denotes the number of valid UV locations.

We then combine sparse vertex-aligned features and dense UV features to form a unified multi-resolution representation.
While UV densification introduces fine-grained spatial detail, it does not explicitly enforce global appearance consistency across multiple observations. Therefore, we perform a second-stage cross-modal alignment to further refine the fused representation using the global appearance tokens:
\begin{equation} 
\mathbf{T}_{d} \gets \operatorname{\textsc{CrossAtt}} \left( \left[ \mathbf{T}_{\text{uv}}, \mathbf{T}_{s} \right], \mathbf{Z}^{\text{app}} \right)
\end{equation}

This hierarchical alignment progressively transfers identity-consistent appearance information from global image observations to both sparse structural anchors and dense surface samples. As a result, the refined feature set $\mathbf{T}_{d}$ captures semantically stable coarse geometry together with high-frequency surface detail, yielding an efficient and expressive representation for canonical 3D Gaussian field decoding.

\noindent\textbf{Canonical 3D Gaussian Field Decoding.}
The refined feature set $\mathbf{T}_d$ is decoded into a canonical 3D Gaussian head representation using a feed-forward decoder $\mathcal{D}{\text{static}}$. For each Gaussian primitive, the decoder predicts both geometric and appearance attributes:
\begin{equation} %
G=\{f_n,o_n,c_n,s_n,r_n\}_{n=1}^{M+5023} = \mathcal{D}_{\text{static}}(\mathbf{T}_d)
\end{equation} 
here, $\mathbf{f}_n \in \mathbb{R}^3$ denotes the positional offset relative to the underlying FLAME head template, $o_n \in \mathbb{R}$ represents opacity, $\mathbf{c}_n \in \mathbb{R}^3$ encodes color, $\mathbf{s}_n \in \mathbb{R}^3$ parameterizes anisotropic scaling, and $\mathbf{r}_n \in \mathbb{R}^4$ denotes rotation represented as a quaternion.
The resulting set of $M{+}5023$ Gaussian primitives defines a viewpoint- and expression-invariant canonical appearance field, which serves as the static structural basis for subsequent dynamic animation.

\subsection{3D Head Avatar Animation}
\label{sec:avatar_animation}

Given the canonical static 3D Gaussian head representation $G$ obtained in Section~\ref{sec:static_canonical_field},
we animate the avatar under arbitrary facial expressions and head poses by combining the FLAME parametric head model~\cite{li2017learning} with a Motion-Aware Refinement Module (MARM).

\noindent\textbf{FLAME-based Mesh Deformation.}
FLAME parameterizes head geometry using pose and expression coefficients.
For each animation frame, the deformation parameters are defined as
\begin{equation}
\boldsymbol{\theta} = \{ \boldsymbol{\theta}^{\text{pose}}, \boldsymbol{\theta}^{\text{exp}} \},
\end{equation}
where $\boldsymbol{\theta}^{\text{pose}}$ models rigid head motion and jaw articulation, and
$\boldsymbol{\theta}^{\text{exp}}$ controls non-rigid facial expressions.
Given the canonical FLAME template mesh $\mathbf{V}_0$, the posed mesh is obtained via linear blend skinning (LBS) with corrective blendshapes:
\begin{equation}
\mathbf{V}(\boldsymbol{\theta}) = \mathrm{LBS}(\mathbf{V}_0, \boldsymbol{\theta})
\end{equation}

Following FLAME deformation, we propagate the vertex-level motion to a dense set of surface points
using the same UV rasterization strategy as in the canonical construction.
Specifically, the deformed mesh $\mathbf{V}(\boldsymbol{\theta})$ is unwrapped into the UV domain,
yielding a set of valid UV pixels $\Omega_{\text{uv}}$ with cardinality $M$.
Each UV pixel $(u,v) \in \Omega_{\text{uv}}$ is associated with barycentric weights
$\boldsymbol{\alpha}(u,v) = (\alpha,\beta,\gamma)$ and vertex indices $(i,j,k)$ on the deformed mesh.
The corresponding dense surface position is computed as:
\begin{equation}
\tilde{\mathbf{x}}_n
=
\alpha\,\mathbf{V}_i(\boldsymbol{\theta})
+
\beta\,\mathbf{V}_j(\boldsymbol{\theta})
+
\gamma\,\mathbf{V}_k(\boldsymbol{\theta}),
\quad n = 1,\dots,M.
\end{equation}

In addition to the densified surface points, we also retain the original deformed FLAME vertices
$\{ \mathbf{V}_m(\boldsymbol{\theta}) \}_{m=1}^{5023}$ to ensure geometric consistency.
By concatenating both sets, we obtain a pose-dependent coarse surface representation with
$M{+}5023$ points:
\begin{equation}
\tilde{\mathbf{X}}(\boldsymbol{\theta})
=
\left[
\{ \tilde{\mathbf{x}}_n \}_{n=1}^{M}
,
\{ \mathbf{V}_m(\boldsymbol{\theta}) \}_{m=1}^{5023}
\right]
\end{equation}

\noindent\textbf{Gaussian Position Update.}
Each Gaussian primitive is anchored to a corresponding point in
$\tilde{\mathbf{X}}(\boldsymbol{\theta})$.
During animation, FLAME provides coarse, physically consistent surface motion,
while the learned canonical Gaussian offsets are preserved to retain identity-specific geometry.
Consequently, the Gaussians follow pose- and expression-driven deformation while maintaining personalized shape and fine-scale details.

\noindent\textbf{Motion-Aware Refinement.}
While FLAME provides physically plausible coarse deformation, it cannot fully capture identity-specific nonlinear motion patterns, such as subtle muscle dynamics or personalized expression styles.
To compensate for these limitations, we introduce a MARM that predicts residual Gaussian updates conditioned on the current animation state.
Specifically, we learn a motion refinement network $\mathcal{R}{\text{motion}}$ that takes as input the positional encoding of canonical FLAME vertices, the expression and pose parameters $\boldsymbol{\theta}$, and the camera parameters $\boldsymbol{\theta}^{\text{cam}}$.
The module predicts residual Gaussian attribute offsets:
\begin{equation}
\Delta \mathbf{g} \gets \mathcal{R}{\text{motion}}(\mathrm{PE}(\mathbf{V}_0), \boldsymbol{\theta}, \boldsymbol{\theta}^{\text{cam}})
\end{equation}

Here, $\Delta \mathbf{g}$ models motion-dependent corrections beyond template-driven deformation, allowing the system to refine geometry and appearance in a view-aware manner. This design avoids directly regressing full Gaussian parameters, instead focusing on compact residual updates that preserve canonical identity while adapting to dynamic conditions.
The final animated Gaussian field is obtained by applying the predicted residuals to the FLAME-deformed canonical representation:
\begin{equation}
G(\boldsymbol{\theta}) = G' \oplus \Delta \mathbf{g}
\end{equation}
where $G'$ denotes the FLAME-driven coarse Gaussian field and $\oplus$ represents attribute-wise updates.

\section{Experiments}

\subsection{Experimental Settings}
\label{sec:implementation_details}

\noindent\textbf{Implementation Details.}
Our framework is implemented in PyTorch and optimized using Adam with a learning rate of $4.0 \times 10^{-5}$ for $300{,}000$ iterations. The DINOv3~\cite{simeoni2025dinov3} feature backbone is frozen during training, while all other components are optimized end-to-end. For each batch, we randomly sample 1$\sim$8 frames from a monocular video to form the input set. The sparse-to-dense cross-modal alignment module and the motion-aware refinement module are implemented using Transformer architectures. We use the GAGAvatar tracker~\cite{chu2024generalizable} to extract camera-pose and facial-expression conditions, which are used to control the viewpoint and expression of the reconstructed avatar during animation. More implementation details are provided in the supplementary material.

\noindent\textbf{Baselines.}
In the single-reference setting, we compare with Portrait4D-v2~\cite{deng2024portrait4dv2}, Real3D-Portrait~\cite{ye2024real3d}, GAGAvatar~\cite{chu2024generalizable}, and LAM~\cite{he2025lam}. In the few-reference setting, we further compare with the optimization-based methods FlashAvatar~\cite{xiang2024flashavatar}, GHA~\cite{xu2024gaussian}, and GaussianAvatars~\cite{qian2024gaussianavatars}, as well as the feed-forward approaches GAPAvatar~\cite{chu2024gpavatar} and FastAvatar~\cite{wu2025fastavatar}. We do not include Avat3r~\cite{kirschstein2025avat3r} and FastGHA~\cite{ji2026fastgha} in the main comparison because their official implementations are not publicly available.

\noindent\textbf{Datasets and Evaluation.}
We train the model on the VFHQ dataset~\cite{xie2022vfhq}, which contains 15,204 monocular video clips (about 3M frames). For each extracted frame, we detect the facial region, enlarge the bounding box to include sufficient contextual information, and crop the region of interest. To enhance data usability, we perform camera pose estimation and FLAME parameter tracking for each frame, following a pipeline similar to GAGAvatar~\cite{chu2024generalizable}.
All cropped images are resized to $512 \times 512$ pixels for consistency. In addition, background removal is applied to separate the subject from the background.
On the VFHQ test set, 8 expressions are randomly selected as inputs, while the remaining frames are used for evaluation. We further evaluate on the NeRSemble dataset~\cite{kirschstein2023nersemble} under two settings: \textit{novel view synthesis} (8 input views and 8 unseen views) and \textit{novel expression synthesis} (16 input expressions with unseen poses and expressions). Rendering quality is measured using PSNR, SSIM~\cite{wang2004image}, and LPIPS~\cite{zhang2018unreasonable}, while identity consistency and motion accuracy are evaluated using CSIM~\cite{deng2019arcface}, AED, and APD. To provide a more complete evaluation against recent feed-forward avatar reconstruction methods, we also conduct qualitative comparisons with Avat3r~\cite{kirschstein2025avat3r} and FastGHA~\cite{ji2026fastgha} on the Ava-256 dataset~\cite{martinez2024codec} using the same color-calibrated input data.

\begin{figure}[t]
\centering
\includegraphics[width=1.0\linewidth]{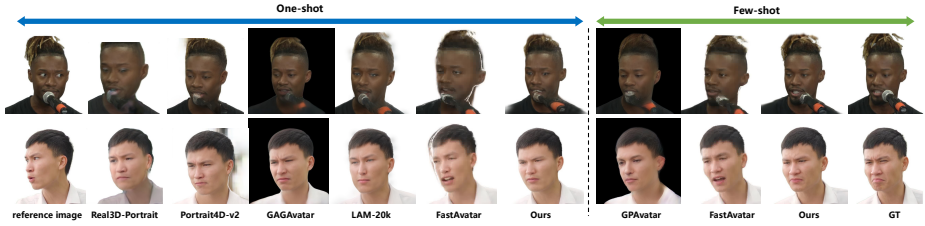}
\caption{\label{fig:novel_exp}Novel expression synthesis. Qualitative comparison of novel expression synthesis on the VFHQ monocular test set. We compare our method with Real3D-Portrait~\cite{ye2024real3d}, Portrait4D-v2~\cite{deng2024portrait4dv2}, GAGAvatar~\cite{chu2024generalizable}, LAM~\cite{he2025lam}, GPAvatar~\cite{chu2024gpavatar}, and FastAvatar~\cite{wu2025fastavatar}.}
\end{figure}

\begin{table}[t]
\centering
\caption{\label{tab:novel-expression}Quantitative comparison on the VFHQ dataset under the novel expression setting. $\uparrow$ indicates higher is better, $\downarrow$ indicates lower is better. We mark the \textbf{best} and \underline{second-best} results.}
\resizebox{\linewidth}{!}{%
\begin{threeparttable}
\begin{tabular}{l|c|cccccc|ccc}
\toprule
\multirow{2}{*}{{Method}} & \multirow{2}{*}{{Setting}}&
\multicolumn{6}{c|}{{Novel Expressions}} & 
\multicolumn{2}{c}{{Efficiency}} \\ 
 \cmidrule(lr){3-5} \cmidrule(lr){6-8} \cmidrule(lr){9-10}
 & 
  & {PSNR}$\uparrow$
 & {SSIM}$\uparrow$ 
 & {LPIPS}$\downarrow$ 
 & {CSIM}$\uparrow$
 & {AED}$\downarrow$ 
 & {APD}$\downarrow$
 & {Creation in}$\downarrow$ 
 & {FPS}$\uparrow$
 \\ 
\midrule
Real3DPortrait~\cite{ye2024real3d} & \multirow{6}{*}{{One-shot}}& 20.88 & 0.780 & 0.154 & 0.750 & 0.150 & 0.268   & 3.5$s$ & 15 \\
Portrait4D-v2~\cite{deng2024portrait4dv2} & & 21.34 & 0.794 & 0.144 & 0.717 & 0.117 & 0.187  & 2.9$s$ & 11\\
GAGAvatar~\cite{chu2024generalizable}   &   & 21.83 & 0.818 & 0.128 & 0.816 & 0.111 & 0.135   & 1.6$s$ & {63} \\
LAM~\cite{he2025lam}   &   & 22.65 & 0.829 & 0.109 & 0.822 & 0.102 & 0.134  & \underline{1.1$s$} & {219} \\
FastAvatar~\cite{wu2025fastavatar}   &   & {17.85} & {0.813} & {0.167} & {0.679} & {0.136} & {0.328}& {2.6$s$} & \underline{339} \\

Ours   &   & {21.82} & {0.843} & {0.108} & {0.817} & {0.109} & {0.149}& {1.3$s$} & {31} \\

\midrule
{GPAvatar}~\cite{chu2024gpavatar} & \multirow{4}{*}{{Few-shot}}
& {22.91} & {0.795} & {0.154}
& {0.765} & {0.138} & {0.189}
& \textbf{0.7$s$} & {5} \\

FastAvatar~\cite{wu2025fastavatar}   &   & {18.12} & {0.819} & {0.153} & {0.781} & {0.116} & {0.321}& {12.2$s$} & {97} \\

{Ours(fast)} & 
& \underline{23.20} & \underline{0.862} & \underline{0.088}
& \underline{0.852} & \underline{0.084} & \underline{0.117}
& {2.1$s$} & \textbf{468} \\
{Ours} & 
& \textbf{23.35} & \textbf{0.864} & \textbf{0.081}
& \textbf{0.861} & \textbf{0.079} & \textbf{0.114}
& {2.1$s$} & {31} \\
\midrule
\end{tabular}
\end{threeparttable}
}
\end{table}

\subsection{Head Avatar Reconstruction}
\noindent\textbf{Quantitative and Qualitative Results on VFHQ.}
Table~\ref{tab:novel-expression} reports quantitative results on the VFHQ monocular test set under the novel-expression setting. 
Compared with one-shot feed-forward reconstruction methods, a key advantage of our approach lies in its ability to flexibly incorporate multiple reference images for incremental reconstruction. 
Benefiting from this capability, under the few-shot setting our method achieves the best performance across all image quality metrics (PSNR, SSIM, and LPIPS) for novel-expression rendering, while also obtaining higher identity consistency (CSIM) and better motion fidelity (AED and APD).
Efficiency comparisons further highlight the practicality of our approach. 
The fast variant (without the motion refinement module) achieves real-time rendering at 468 FPS, significantly surpassing all baselines while maintaining competitive visual quality. 
Although the full model prioritizes rendering fidelity, it remains computationally efficient and does not require per-subject optimization, resulting in only moderate avatar creation time.
Qualitative comparisons are shown in Fig.~\ref{fig:novel_exp}. 
\begin{figure}[!t]
\centering
\includegraphics[width=1.0\linewidth]{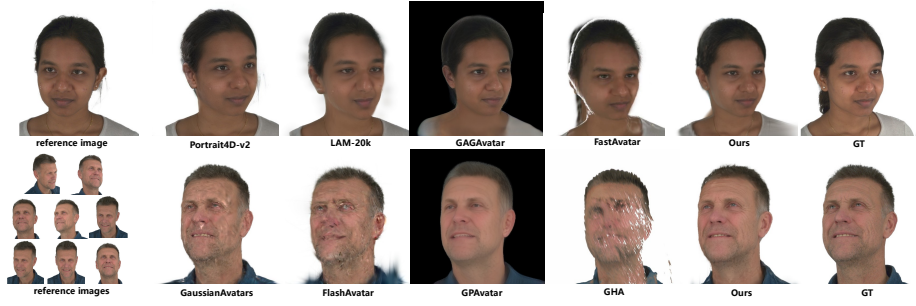}
\caption{\label{fig:novel_view}Novel View Synthesis. Qualitative comparison of novel view synthesis results on the NeRsemble multi-view subset. For the single-reference setting, we compare our method with Portrait4D-v2~\cite{deng2024portrait4dv2}, GAGAvatar~\cite{chu2024generalizable}, LAM~\cite{he2025lam}, and FastAvatar~\cite{wu2025fastavatar}. For the multi-reference setting, we further compare against GaussianAvatars~\cite{qian2024gaussianavatars}, FlashAvatar~\cite{xiang2024flashavatar}, GHA~\cite{xu2024gaussian}, and GPAvatar~\cite{chu2024gpavatar}.}
\end{figure}

\begin{table}[!t]
\centering
\caption{\label{tab:novel-view}Quantitative comparison on the NeRSemble dataset under novel-view and novel-expression settings.
$\uparrow$ indicates higher is better, $\downarrow$ indicates lower is better. We mark the \textbf{best} and \underline{second-best} results.}

\resizebox{\linewidth}{!}{%
\begin{threeparttable}
\begin{tabular}{l|c|ccc|ccc}
\toprule

\multirow{2}{*}{{Method}} & \multirow{2}{*}{{Setting}}&
\multicolumn{3}{c|}{{Novel Views}} & 
\multicolumn{3}{c}{{Novel Expressions}} \\ 
 \cmidrule(lr){3-5} \cmidrule(lr){6-8} 
 & 
 & {LPIPS}$\downarrow$ 
 & {SSIM}$\uparrow$ 
 & {PSNR}$\uparrow$
 & {LPIPS}$\downarrow$
 & {SSIM}$\uparrow$ 
 & {PSNR}$\uparrow$
 \\ 
\midrule

{Real3DPortrait}~\cite{ye2024real3d} & \multirow{6}{*}{{One-shot}}
& {0.197} & {0.785} & {16.22}
& {0.165} & {0.821} & {17.48}
 \\

{Portrait4D-v2}~\cite{deng2024portrait4dv2} &
& {0.172} & {0.797} & {16.81}
& {0.152} & {0.814} & {18.24}
 \\

{GAGAvatar}~\cite{chu2024generalizable} &
& {0.129} & {0.833} & \textbf{22.52}
& \underline{0.095} & \underline{0.857} & \textbf{25.87}
 \\

{LAM-20K}~\cite{he2025lam} &
& {0.175} & {0.819} & {16.43}
& {0.122} & {0.834} & {20.55}
\\
{FastAvatar}~\cite{wu2025fastavatar} &
& {0.232} & {0.800} & {14.78}
& {0.185} & {0.821} & {19.41}
\\

{Ours} & 
& \underline{0.121} & \underline{0.839} & {19.18}
& {0.106} & {0.851} & {20.23}\\
\midrule
{FlashAvatar}~\cite{xiang2024flashavatar} & \multirow{3}{*}{{Train From Scratch}}
& {0.209} & {0.785} & {17.84}
& {0.221} & {0.764} & {16.94}
 \\
{GHA}~\cite{xu2024gaussian} &
& {0.269} & {0.722} & {13.93}
& {-} & {-} & {-} \\

{GaussianAvatars}~\cite{qian2024gaussianavatars} &
& {0.164} & {0.813} & {17.99}
& {0.178} & {0.822} & {17.56} \\

\midrule
{GPAvatar}~\cite{chu2024gpavatar} & \multirow{3}{*}{{Few-shot}}
& {0.163} & {0.822} & \underline{22.26}
& {0.154} & {0.829} & {22.58}\\

{FastAvatar}~\cite{wu2025fastavatar} &
& {0.158} & {0.824} & {20.11}
& {0.135} & {0.845} & {22.49}
\\

{Ours} & 
& \textbf{0.098} & \textbf{0.858} & {21.95}
& \textbf{0.075} & \textbf{0.881} & \underline{24.08}\\
\midrule
\end{tabular}
\end{threeparttable}
}
\end{table}

\noindent\textbf{Quantitative and Qualitative Results on NeRSemble.}
It is worth noting that our model is trained exclusively on the VFHQ monocular dataset. The NeRSemble multi-view dataset is not used during training and serves solely for evaluation.
Table~\ref{tab:novel-view} presents quantitative comparisons on the NeRSemble multi-view dataset under both novel-view and novel-expression settings. 
In the one-shot scenario, our method achieves competitive performance compared with existing feed-forward approaches, outperforming most baselines in perceptual quality metrics while maintaining stable rendering quality across unseen viewpoints and expressions. 
More importantly, when multiple reference images are available, our framework demonstrates clear advantages. Under the few-shot setting, our method achieves the best performance in LPIPS and SSIM for both novel-view and novel-expression synthesis, indicating superior perceptual fidelity and structural consistency. Although GAGAvatar reports slightly higher PSNR in the one-shot setting, our method provides a better overall balance between perceptual quality and geometric consistency, particularly when leveraging multiple input views.
Figure~\ref{fig:novel_view} further presents qualitative comparisons for novel-view synthesis. Compared with Gaussian-based or train-from-scratch baselines, our method produces sharper facial structures, fewer view-dependent artifacts, and more coherent multi-view appearance, whereas competing approaches often exhibit blur or geometric inconsistencies.

\begin{figure}[!t]
\centering
\includegraphics[width=1.0\linewidth]{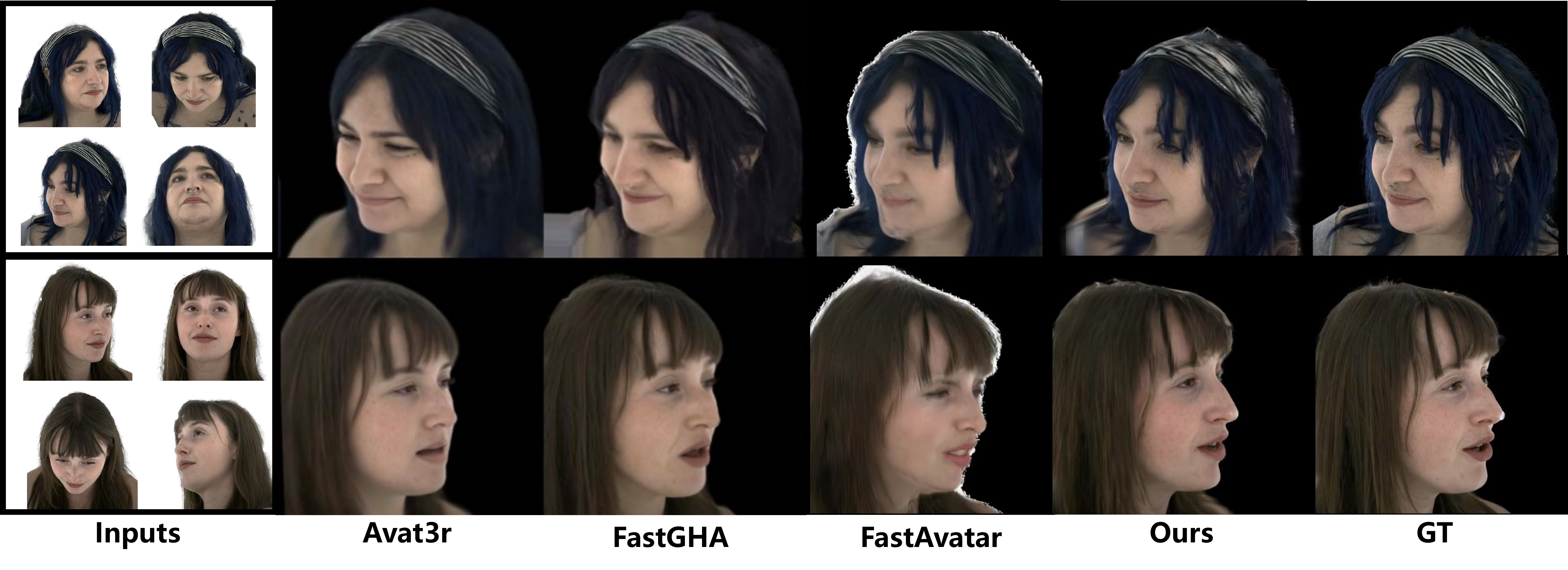}
\caption{
\textbf{Qualitative comparisons with FastGHA and Avat3r.}
Using the same color-calibrated Ava-256 inputs~\cite{martinez2024codec}, FFAvatar preserves sharper facial details and more stable identity consistency under comparable reference-input settings.
}
\label{fig:supp_ava256}
\end{figure}

\noindent\textbf{Qualitative Comparisons on Ava-256.}
As discussed in the main paper, we do not include Avat3r~\cite{kirschstein2025avat3r} and FastGHA~\cite{ji2026fastgha} in the main quantitative comparison because their official implementations are not publicly available.
For completeness, Fig.~\ref{fig:supp_ava256} provides additional qualitative comparisons on the Ava-256 dataset~\cite{martinez2024codec}.
Following the meta-review suggestion, we conduct the comparison using the same color-calibrated input data for all methods, ensuring a consistent input setting.
These results complement the main evaluation and provide a visual reference for recent feed-forward avatar reconstruction methods.
Compared with FastGHA and Avat3r, FFAvatar better preserves fine facial details and maintains more stable identity consistency across the rendered views.

\subsection{Ablation Studies}

\begin{figure}[!t]
\centering
\includegraphics[width=1.0\linewidth]{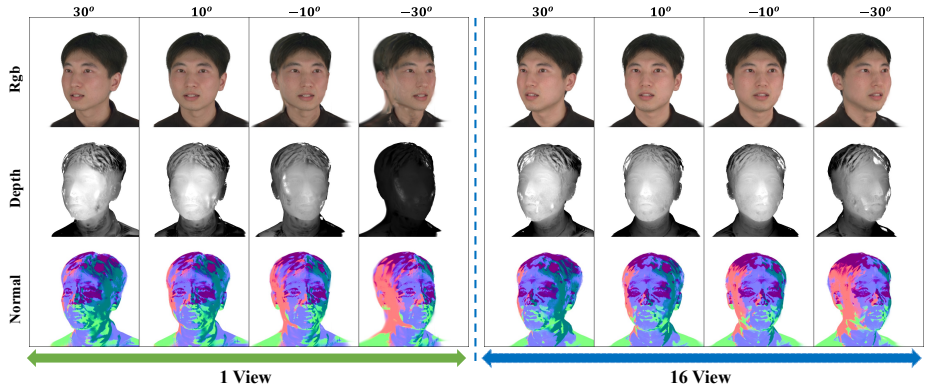}
\caption{\label{fig:multi_inputs}Multi-view rendering results of our method under different input conditions: single-image input (left) and 16-image input (right). The rendered viewpoints correspond to yaw angles of $30^\circ$, $10^\circ$, $-10^\circ$, and $-30^\circ$.  From top to bottom, we show the RGB renderings, depth maps, and surface normal maps produced by 3D Gaussian splatting.}
\end{figure}
\noindent\textbf{Incremental Reconstruction.}
FFAvatar supports incremental reconstruction from one or more reference portrait images with diverse expressions and viewpoints.
As shown in Table~\ref{tab:multi_inputs}, reconstruction quality improves as the number of input frames increases.
The most notable gain is observed when increasing the inputs from 1 to 8 frames, suggesting that additional multi-view and multi-expression cues are particularly beneficial for identity stabilization and motion alignment.
Further increasing the number of inputs to 16 or 32 frames brings smaller but consistent improvements, indicating that FFAvatar can effectively integrate additional observations.
Figure~\ref{fig:multi_inputs} provides qualitative comparisons.
With only a single input image, the reconstructed avatar already preserves plausible geometry and appearance, although minor artifacts and view-dependent inconsistencies may appear under large pose changes.
When more input images are provided, such as 16 frames, facial structures become sharper, view transitions are smoother, and identity preservation becomes more stable across viewpoints.

\begin{table}[!t]
\centering
\caption{\label{tab:multi_inputs}Ablation study on the number of input images under the novel expression setting on the VFHQ dataset. We analyze reconstruction quality, motion fidelity, identity consistency, and creation time as the number of input frames increases.}
\resizebox{\linewidth}{!}{%
\begin{threeparttable}
\begin{tabular}{l|c|cccccc|c}
\toprule
\multirow{2}{*}{{Method}} & \multirow{2}{*}{{Inputs}} & \multicolumn{6}{c|}{{Novel Expressions}} & \multicolumn{1}{c}{{Efficiency}} \\
\cmidrule(lr){3-5} \cmidrule(lr){6-8} \cmidrule(lr){9-9}
& & {PSNR}$\uparrow$ & {SSIM}$\uparrow$ & {LPIPS}$\downarrow$ & {CSIM}$\uparrow$ & {AED}$\downarrow$ & {APD}$\downarrow$ & {Creation in}$\downarrow$ \\
\midrule
\multirow{5}{*}{Ours} & 1 & {21.82} & {0.843} & {0.108} & {0.817} & {0.109} & {0.149} & \textbf{1.3$s$} \\
& 4 & {22.75} & {0.858} & {0.091} & {0.852} & {0.094} & {0.119} & {1.7$s$} \\
& 8 & {23.35} & {0.864} & {0.081} & {0.861} & {0.079} & {0.114} & {2.1$s$} \\

& 16 & {23.36} & {0.866} & {0.080} & {0.872} & {0.078} & \textbf{0.110} & {4.3$s$} \\
& 32 & \textbf{23.38} & \textbf{0.867} & \textbf{0.077} & \textbf{0.874} & \textbf{0.078} & {0.111} & {11.6$s$} \\

\midrule
\end{tabular}
\end{threeparttable}
}
\end{table}

\begin{figure}[!t]
\centering
\includegraphics[width=1.0\linewidth]{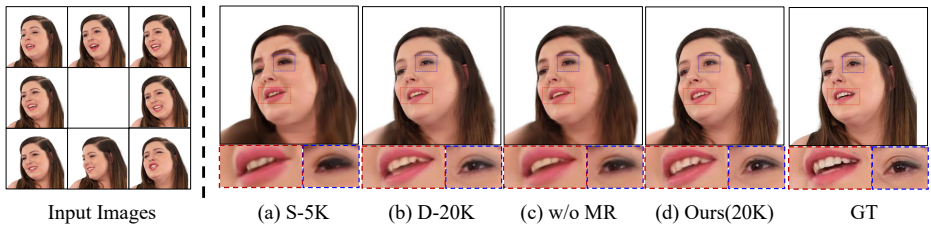}
\caption{\label{fig:ablation_study}Ablation study of key design components. 
(a) Optimization using the original FLAME vertices as Gaussian primitives results in insufficient texture detail in several regions; 
(b) direct optimization of dense point clouds leads to suboptimal deformation, with some primitives not well optimized; 
(c) removing the motion refinement module causes inaccurate dynamic motion representation; 
(d) final results of the full model, producing the most detailed appearance and accurate motion.}
\end{figure}

\noindent\textbf{Sparse-to-Dense Learning Paradigm.} We further evaluate the effectiveness of the proposed sparse-to-dense appearance modeling strategy.
Our design first learns coarse yet semantically stable appearance features at the sparse FLAME vertex level, 
and then densifies the representation in UV space to capture high-frequency geometric and photometric details.
Without such hierarchical modeling, directly optimizing sparse primitives leads to insufficient representation 
capacity for fine structures (e.g., hair strands and subtle facial textures), whereas directly optimizing dense 
Gaussian primitives significantly increases training cost and memory consumption, and may result in suboptimal 
deformation due to insufficiently constrained optimization.
As shown in Table~\ref{tab:sparse-to-dense}, the sparse-only model (S-5K) exhibits clear performance degradation 
across all quality and motion metrics, indicating limited representational capability.
In contrast, the dense-only optimization (D-20K, D-64K) improves certain reconstruction metrics but incurs 
substantially higher training time and GPU memory usage, and even becomes infeasible at higher resolutions 
due to memory overflow.
Qualitative comparisons in Fig.~\ref{fig:ablation_study} further support these findings.
Direct sparse optimization produces overly smooth appearance with missing high-frequency details, 
while direct dense optimization leads to unstable or imperfect deformation in certain regions. In contrast, the proposed sparse-to-dense paradigm enables both detailed appearance modeling and an efficient training scheme, achieving a more favorable trade-off between reconstruction accuracy and computational efficiency.

\begin{table}[!t]
\centering
\caption{\label{tab:sparse-to-dense}Effect of the sparse-to-dense generation strategy. 
S denotes optimization using the original FLAME vertex set (5,023 points), 
D denotes direct optimization of dense point clouds, and S2D represents the proposed sparse-to-dense generation process. 
``--'' indicates the number of Gaussian primitives. 
OOM indicates out-of-memory under a 40\,GB GPU memory constraint. Time refers to the wall-clock time required for 1,000 training iterations.}
\resizebox{\linewidth}{!}{%
\begin{threeparttable}
\begin{tabular}{l|c|cccccc|cc}
\toprule
\multirow{2}{*}{{Method}} & \multirow{2}{*}{{UV Resolution}} & \multicolumn{6}{c|}{{Novel Expressions}} & \multicolumn{2}{c}{{Train}} \\
\cmidrule(lr){3-5} \cmidrule(lr){6-8} \cmidrule(lr){9-10}
& & {PSNR}$\uparrow$ & {SSIM}$\uparrow$ & {LPIPS}$\downarrow$ & {CSIM}$\uparrow$ & {AED}$\downarrow$ & {APD}$\downarrow$ & {Time}$\downarrow$ & {Memory}$\downarrow$ \\
\midrule

S-5K& - & {19.69} & {0.837} & {0.174} & {0.689} & {0.131} & {0.141} & {40min} & 22.5GB \\
S2D-20K& 128 & {23.35} & {0.864} & {0.081} & {0.861} & {0.079} & {0.114} & {47min} &25.8GB \\
D-20K& 128 & {23.41} & {0.869} & {0.097} & {0.858} & {0.087} & {0.115} & {120min}&38.2GB \\
\midrule
S2D-64K& 256 & \textbf{23.42} & \textbf{0.871} & \textbf{0.077} & \textbf{0.866} & \textbf{0.078} & \textbf{0.111} & {72min}&37.1GB \\
D-64K& 256 & {-} & {-} & {-} & {-} & {-} & {-} & {-}&OOM \\
\midrule
\end{tabular}
\end{threeparttable}
}
\end{table}

\noindent\textbf{Motion Refinement Module.} We evaluate the effectiveness of the motion refinement module, which enhances dynamic deformation beyond the FLAME template. 
As shown in Table~\ref{tab:novel-expression}, enabling motion refinement (Ours) consistently improves perceptual quality, identity consistency, and motion accuracy compared to the fast variant without refinement (Ours(fast)), while maintaining similar creation time. Fig.~\ref{fig:ablation_study}(c) further shows that removing this module leads to less accurate and oversimplified motion, whereas the full model produces more natural and detailed dynamic behavior.

\section{Conclusion}

We present \textbf{FFAvatar}, an incremental feed-forward framework for efficient and high-fidelity reconstruction of 4D avatar heads from sparse and heterogeneous image inputs. 
Through an alternating attention mechanism, our method flexibly disentangles identity information from expression and viewpoint across a variable number of reference images, enabling the learning of a stable canonical 3D appearance and ensuring strong cross-view and cross-expression consistency. 
We propose a sparse-to-dense learning paradigm, where sparse point representations first capture the coarse geometry of the 3D avatar and are subsequently densified in UV space to recover fine-grained geometric and texture details, achieving a better balance between fidelity and efficiency.
In addition, a motion refinement module is incorporated to further enhance the realism of dynamic expressions. 
Overall, FFAvatar provides a flexible, efficient, and robust solution for reconstructing controllable 4D avatars from one or more portrait images.

\section*{Acknowledgements}
This work was supported by the National Natural Science Foundation of China under Grant 62406120, the Hubei Provincial Natural Science Foundation of China under Grant No. 2026AFB533 and the CCF-Zhipu Large Model Innovation Fund (NO.CCF-Zhipu202411).

\bibliographystyle{splncs04}
\bibliography{main}

@String(CVPR  = {IEEE Conf. Comput. Vis. Pattern Recog.})

@String(TOG   = {ACM Trans. Graph.})

@String(CVPR  = {CVPR})

@String(TOG   = {ACM TOG})

@article{nerf,
  title={Nerf: Representing scenes as neural radiance fields for view synthesis},
  author={Mildenhall, Ben and Srinivasan, Pratul P and Tancik, Matthew and Barron, Jonathan T and Ramamoorthi, Ravi and Ng, Ren},
  journal={Communications of the ACM},
  volume={65},
  number={1},
  pages={99--106},
  year={2021},
  publisher={ACM New York, NY, USA}
}

@article{instant,
  title={Instant neural graphics primitives with a multiresolution hash encoding},
  author={M{\"u}ller, Thomas and Evans, Alex and Schied, Christoph and Keller, Alexander},
  journal={ACM transactions on graphics (TOG)},
  volume={41},
  number={4},
  pages={1--15},
  year={2022},
  publisher={ACM New York, NY, USA}
}

@inproceedings{athar2022rignerf,
  title={Rignerf: Fully controllable neural 3d portraits},
  author={Athar, ShahRukh and Xu, Zexiang and Sunkavalli, Kalyan and Shechtman, Eli and Shu, Zhixin},
  booktitle={Proceedings of the IEEE/CVF conference on Computer Vision and Pattern Recognition},
  pages={20364--20373},
  year={2022}
}

@inproceedings{gafni2021dynamic,
  title={Dynamic neural radiance fields for monocular 4d facial avatar reconstruction},
  author={Gafni, Guy and Thies, Justus and Zollhofer, Michael and Nie{\ss}ner, Matthias},
  booktitle={Proceedings of the IEEE/CVF conference on computer vision and pattern recognition},
  pages={8649--8658},
  year={2021}
}

@inproceedings{hong2022headnerf,
  title={Headnerf: A real-time nerf-based parametric head model},
  author={Hong, Yang and Peng, Bo and Xiao, Haiyao and Liu, Ligang and Zhang, Juyong},
  booktitle={Proceedings of the IEEE/CVF Conference on Computer Vision and Pattern Recognition},
  pages={20374--20384},
  year={2022}
}

@inproceedings{kania2022conerf,
  title={Conerf: Controllable neural radiance fields},
  author={Kania, Kacper and Yi, Kwang Moo and Kowalski, Marek and Trzci{\'n}ski, Tomasz and Tagliasacchi, Andrea},
  booktitle={Proceedings of the IEEE/CVF Conference on Computer Vision and Pattern Recognition},
  pages={18623--18632},
  year={2022}
}

@inproceedings{park2021nerfies,
  title={Nerfies: Deformable neural radiance fields},
  author={Park, Keunhong and Sinha, Utkarsh and Barron, Jonathan T and Bouaziz, Sofien and Goldman, Dan B and Seitz, Steven M and Martin-Brualla, Ricardo},
  booktitle={Proceedings of the IEEE/CVF international conference on computer vision},
  pages={5865--5874},
  year={2021}
}

@inproceedings{xiang2024flashavatar,
  title={Flashavatar: High-fidelity head avatar with efficient gaussian embedding},
  author={Xiang, Jun and Gao, Xuan and Guo, Yudong and Zhang, Juyong},
  booktitle={Proceedings of the IEEE/CVF Conference on Computer Vision and Pattern Recognition},
  pages={1802--1812},
  year={2024}
}

@inproceedings{qian2024gaussianavatars,
  title={Gaussianavatars: Photorealistic head avatars with rigged 3d gaussians},
  author={Qian, Shenhan and Kirschstein, Tobias and Schoneveld, Liam and Davoli, Davide and Giebenhain, Simon and Nie{\ss}ner, Matthias},
  booktitle={Proceedings of the IEEE/CVF Conference on Computer Vision and Pattern Recognition},
  pages={20299--20309},
  year={2024}
}

@inproceedings{guo2021ad,
  title={Ad-nerf: Audio driven neural radiance fields for talking head synthesis},
  author={Guo, Yudong and Chen, Keyu and Liang, Sen and Liu, Yong-Jin and Bao, Hujun and Zhang, Juyong},
  booktitle={Proceedings of the IEEE/CVF international conference on computer vision},
  pages={5784--5794},
  year={2021}
}

@inproceedings{li2023efficient,
  title={Efficient region-aware neural radiance fields for high-fidelity talking portrait synthesis},
  author={Li, Jiahe and Zhang, Jiawei and Bai, Xiao and Zhou, Jun and Gu, Lin},
  booktitle={Proceedings of the IEEE/CVF International Conference on Computer Vision},
  pages={7568--7578},
  year={2023}
}

@article{aneja2024gaussianspeech,
  title={Gaussianspeech: Audio-driven gaussian avatars},
  author={Aneja, Shivangi and Sevastopolsky, Artem and Kirschstein, Tobias and Thies, Justus and Dai, Angela and Nie{\ss}ner, Matthias},
  journal={arXiv preprint arXiv:2411.18675},
  year={2024}
}

@inproceedings{cho2024gaussiantalker,
  title={Gaussiantalker: Real-time talking head synthesis with 3d gaussian splatting},
  author={Cho, Kyusun and Lee, Joungbin and Yoon, Heeji and Hong, Yeobin and Ko, Jaehoon and Ahn, Sangjun and Kim, Seungryong},
  booktitle={Proceedings of the 32nd ACM International Conference on Multimedia},
  pages={10985--10994},
  year={2024}
}

@inproceedings{peng2024synctalk,
  title={Synctalk: The devil is in the synchronization for talking head synthesis},
  author={Peng, Ziqiao and Hu, Wentao and Shi, Yue and Zhu, Xiangyu and Zhang, Xiaomei and Zhao, Hao and He, Jun and Liu, Hongyan and Fan, Zhaoxin},
  booktitle={Proceedings of the IEEE/CVF Conference on Computer Vision and Pattern Recognition},
  pages={666--676},
  year={2024}
}

@inproceedings{li2024talkinggaussian,
  title={Talkinggaussian: Structure-persistent 3d talking head synthesis via gaussian splatting},
  author={Li, Jiahe and Zhang, Jiawei and Bai, Xiao and Zheng, Jin and Ning, Xin and Zhou, Jun and Gu, Lin},
  booktitle={European Conference on Computer Vision},
  pages={127--145},
  year={2024},
  organization={Springer}
}

@inproceedings{he2025lam,
  title={LAM: large avatar model for one-shot animatable gaussian head},
  author={He, Yisheng and Gu, Xiaodong and Ye, Xiaodan and Xu, Chao and Zhao, Zhengyi and Dong, Yuan and Yuan, Weihao and Dong, Zilong and Bo, Liefeng},
  booktitle={Proceedings of the Special Interest Group on Computer Graphics and Interactive Techniques Conference Conference Papers},
  pages={1--13},
  year={2025}
}

@article{chu2024generalizable,
  title={Generalizable and animatable gaussian head avatar},
  author={Chu, Xuangeng and Harada, Tatsuya},
  journal={Advances in Neural Information Processing Systems},
  volume={37},
  pages={57642--57670},
  year={2024}
}

@article{jiang2025uv,
  title={Uv gaussians: Joint learning of mesh deformation and gaussian textures for human avatar modeling},
  author={Jiang, Yujiao and Liao, Qingmin and Li, Xiaoyu and Ma, Li and Zhang, Qi and Zhang, Chaopeng and Lu, Zongqing and Shan, Ying},
  journal={Knowledge-Based Systems},
  volume={320},
  pages={113470},
  year={2025},
  publisher={Elsevier}
}

@inproceedings{kirschstein2024gghead,
  title={Gghead: Fast and generalizable 3d gaussian heads},
  author={Kirschstein, Tobias and Giebenhain, Simon and Tang, Jiapeng and Georgopoulos, Markos and Nie{\ss}ner, Matthias},
  booktitle={SIGGRAPH Asia 2024 Conference Papers},
  pages={1--11},
  year={2024}
}

@article{xu2025vasa,
  title={VASA-3D: Lifelike Audio-Driven Gaussian Head Avatars from a Single Image},
  author={Xu, Sicheng and Chen, Guojun and Yang, Jiaolong and Zhang, Yizhong and Deng, Yu and Lin, Steve and Guo, Baining},
  journal={arXiv preprint arXiv:2512.14677},
  year={2025}
}

@inproceedings{kirschstein2025avat3r,
  title={Avat3r: Large animatable gaussian reconstruction model for high-fidelity 3d head avatars},
  author={Kirschstein, Tobias and Romero, Javier and Sevastopolsky, Artem and Nie{\ss}ner, Matthias and Saito, Shunsuke},
  booktitle={Proceedings of the IEEE/CVF International Conference on Computer Vision},
  pages={12089--12100},
  year={2025}
}

@inproceedings{doukas2021headgan,
  title={Headgan: One-shot neural head synthesis and editing},
  author={Doukas, Michail Christos and Zafeiriou, Stefanos and Sharmanska, Viktoriia},
  booktitle={Proceedings of the IEEE/CVF International conference on Computer Vision},
  pages={14398--14407},
  year={2021}
}

@inproceedings{drobyshev2022megaportraits,
  title={Megaportraits: One-shot megapixel neural head avatars},
  author={Drobyshev, Nikita and Chelishev, Jenya and Khakhulin, Taras and Ivakhnenko, Aleksei and Lempitsky, Victor and Zakharov, Egor},
  booktitle={Proceedings of the 30th ACM International Conference on Multimedia},
  pages={2663--2671},
  year={2022}
}

@inproceedings{khakhulin2022realistic,
  title={Realistic one-shot mesh-based head avatars},
  author={Khakhulin, Taras and Sklyarova, Vanessa and Lempitsky, Victor and Zakharov, Egor},
  booktitle={European Conference on Computer Vision},
  pages={345--362},
  year={2022},
  organization={Springer}
}

@inproceedings{tewari2020stylerig,
  title={Stylerig: Rigging stylegan for 3d control over portrait images},
  author={Tewari, Ayush and Elgharib, Mohamed and Bharaj, Gaurav and Bernard, Florian and Seidel, Hans-Peter and P{\'e}rez, Patrick and Zollhofer, Michael and Theobalt, Christian},
  booktitle={Proceedings of the IEEE/CVF conference on computer vision and pattern recognition},
  pages={6142--6151},
  year={2020}
}

@inproceedings{yao2020mesh,
  title={Mesh guided one-shot face reenactment using graph convolutional networks},
  author={Yao, Guangming and Yuan, Yi and Shao, Tianjia and Zhou, Kun},
  booktitle={Proceedings of the 28th ACM international conference on multimedia},
  pages={1773--1781},
  year={2020}
}

@inproceedings{yang2022face2face,
  title={Face2face $\rho$: Real-time high-resolution one-shot face reenactment},
  author={Yang, Kewei and Chen, Kang and Guo, Daoliang and Zhang, Song-Hai and Guo, Yuan-Chen and Zhang, Weidong},
  booktitle={European conference on computer vision},
  pages={55--71},
  year={2022},
  organization={Springer}
}

@article{zeng2022fnevr,
  title={FNeVR: Neural volume rendering for face animation},
  author={Zeng, Bohan and Liu, Boyu and Li, Hong and Liu, Xuhui and Liu, Jianzhuang and Chen, Dapeng and Peng, Wei and Zhang, Baochang},
  journal={Advances in Neural Information Processing Systems},
  volume={35},
  pages={22451--22462},
  year={2022}
}

@article{feng2021learning,
  title={Learning an animatable detailed 3D face model from in-the-wild images},
  author={Feng, Yao and Feng, Haiwen and Black, Michael J and Bolkart, Timo},
  journal={ACM Transactions on Graphics (ToG)},
  volume={40},
  number={4},
  pages={1--13},
  year={2021},
  publisher={ACM New York, NY, USA}
}

@inproceedings{danvevcek2022emoca,
  title={Emoca: Emotion driven monocular face capture and animation},
  author={Dan{\v{e}}{\v{c}}ek, Radek and Black, Michael J and Bolkart, Timo},
  booktitle={Proceedings of the IEEE/CVF conference on computer vision and pattern recognition},
  pages={20311--20322},
  year={2022}
}

@inproceedings{zielonka2022towards,
  title={Towards metrical reconstruction of human faces},
  author={Zielonka, Wojciech and Bolkart, Timo and Thies, Justus},
  booktitle={European conference on computer vision},
  pages={250--269},
  year={2022},
  organization={Springer}
}

@article{filntisis2022visual,
  title={Visual speech-aware perceptual 3d facial expression reconstruction from videos},
  author={Filntisis, Panagiotis P and Retsinas, George and Paraperas-Papantoniou, Foivos and Katsamanis, Athanasios and Roussos, Anastasios and Maragos, Petros},
  journal={arXiv preprint arXiv:2207.11094},
  year={2022}
}

@article{hong2023lrm,
  title={Lrm: Large reconstruction model for single image to 3d},
  author={Hong, Yicong and Zhang, Kai and Gu, Jiuxiang and Bi, Sai and Zhou, Yang and Liu, Difan and Liu, Feng and Sunkavalli, Kalyan and Bui, Trung and Tan, Hao},
  journal={arXiv preprint arXiv:2311.04400},
  year={2023}
}

@article{ren2024l4gm,
  title={L4gm: Large 4d gaussian reconstruction model},
  author={Ren, Jiawei and Xie, Cheng and Mirzaei, Ashkan and Kreis, Karsten and Liu, Ziwei and Torralba, Antonio and Fidler, Sanja and Kim, Seung Wook and Ling, Huan and others},
  journal={Advances in Neural Information Processing Systems},
  volume={37},
  pages={56828--56858},
  year={2024}
}

@article{liang2024feed,
  title={Feed-forward bullet-time reconstruction of dynamic scenes from monocular videos},
  author={Liang, Hanxue and Ren, Jiawei and Mirzaei, Ashkan and Torralba, Antonio and Liu, Ziwei and Gilitschenski, Igor and Fidler, Sanja and Oztireli, Cengiz and Ling, Huan and Gojcic, Zan and others},
  journal={arXiv preprint arXiv:2412.03526},
  year={2024}
}

@article{yang2024storm,
  title={Storm: Spatio-temporal reconstruction model for large-scale outdoor scenes},
  author={Yang, Jiawei and Huang, Jiahui and Chen, Yuxiao and Wang, Yan and Li, Boyi and You, Yurong and Sharma, Apoorva and Igl, Maximilian and Karkus, Peter and Xu, Danfei and others},
  journal={arXiv preprint arXiv:2501.00602},
  year={2024}
}

@article{qi2025predicting,
  title={Predicting 3D representations for Dynamic Scenes},
  author={Qi, Di and Yang, Tong and Wang, Beining and Zhang, Xiangyu and Zhang, Wenqiang},
  journal={arXiv preprint arXiv:2501.16617},
  year={2025}
}

@inproceedings{bai2023learning,
  title={Learning personalized high quality volumetric head avatars from monocular rgb videos},
  author={Bai, Ziqian and Tan, Feitong and Huang, Zeng and Sarkar, Kripasindhu and Tang, Danhang and Qiu, Di and Meka, Abhimitra and Du, Ruofei and Dou, Mingsong and Orts-Escolano, Sergio and others},
  booktitle={Proceedings of the IEEE/CVF Conference on Computer Vision and Pattern Recognition},
  pages={16890--16900},
  year={2023}
}

@inproceedings{giebenhain2024mononphm,
  title={Mononphm: Dynamic head reconstruction from monocular videos},
  author={Giebenhain, Simon and Kirschstein, Tobias and Georgopoulos, Markos and R{\"u}nz, Martin and Agapito, Lourdes and Nie{\ss}ner, Matthias},
  booktitle={Proceedings of the IEEE/CVF conference on computer vision and pattern recognition},
  pages={10747--10758},
  year={2024}
}

@inproceedings{zielonka2023instant,
  title={Instant volumetric head avatars},
  author={Zielonka, Wojciech and Bolkart, Timo and Thies, Justus},
  booktitle={Proceedings of the IEEE/CVF conference on computer vision and pattern recognition},
  pages={4574--4584},
  year={2023}
}

@inproceedings{zheng2023pointavatar,
  title={Pointavatar: Deformable point-based head avatars from videos},
  author={Zheng, Yufeng and Yifan, Wang and Wetzstein, Gordon and Black, Michael J and Hilliges, Otmar},
  booktitle={Proceedings of the IEEE/CVF conference on computer vision and pattern recognition},
  pages={21057--21067},
  year={2023}
}

@inproceedings{athar2024bridging,
  title={Bridging the gap: Studio-like avatar creation from a monocular phone capture},
  author={Athar, ShahRukh and Saito, Shunsuke and Yang, Zhengyu and Pidhorskyi, Stanislav and Cao, Chen},
  booktitle={European Conference on Computer Vision},
  pages={72--88},
  year={2024},
  organization={Springer}
}

@inproceedings{grassal2022neural,
  title={Neural head avatars from monocular rgb videos},
  author={Grassal, Philip-William and Prinzler, Malte and Leistner, Titus and Rother, Carsten and Nie{\ss}ner, Matthias and Thies, Justus},
  booktitle={Proceedings of the IEEE/CVF conference on computer vision and pattern recognition},
  pages={18653--18664},
  year={2022}
}

@article{ichim2015dynamic,
  title={Dynamic 3D avatar creation from hand-held video input},
  author={Ichim, Alexandru Eugen and Bouaziz, Sofien and Pauly, Mark},
  journal={ACM Transactions on Graphics (ToG)},
  volume={34},
  number={4},
  pages={1--14},
  year={2015},
  publisher={ACM New York, NY, USA}
}

@inproceedings{zheng2022avatar,
  title={Im avatar: Implicit morphable head avatars from videos},
  author={Zheng, Yufeng and Abrevaya, Victoria Fern{\'a}ndez and B{\"u}hler, Marcel C and Chen, Xu and Black, Michael J and Hilliges, Otmar},
  booktitle={Proceedings of the IEEE/CVF conference on computer vision and pattern recognition},
  pages={13545--13555},
  year={2022}
}

@inproceedings{ma2021pixel,
  title={Pixel codec avatars},
  author={Ma, Shugao and Simon, Tomas and Saragih, Jason and Wang, Dawei and Li, Yuecheng and De La Torre, Fernando and Sheikh, Yaser},
  booktitle={Proceedings of the IEEE/CVF Conference on Computer Vision and Pattern Recognition},
  pages={64--73},
  year={2021}
}

@inproceedings{wang2025vggt,
  title={Vggt: Visual geometry grounded transformer},
  author={Wang, Jianyuan and Chen, Minghao and Karaev, Nikita and Vedaldi, Andrea and Rupprecht, Christian and Novotny, David},
  booktitle={Proceedings of the Computer Vision and Pattern Recognition Conference},
  pages={5294--5306},
  year={2025}
}

@article{shen2025fastvggt,
  title={Fastvggt: Training-free acceleration of visual geometry transformer},
  author={Shen, You and Zhang, Zhipeng and Qu, Yansong and Zheng, Xiawu and Ji, Jiayi and Zhang, Shengchuan and Cao, Liujuan},
  journal={arXiv preprint arXiv:2509.02560},
  year={2025}
}

@inproceedings{ma20243d,
  title={3d gaussian blendshapes for head avatar animation},
  author={Ma, Shengjie and Weng, Yanlin and Shao, Tianjia and Zhou, Kun},
  booktitle={ACM SIGGRAPH 2024 Conference Papers},
  pages={1--10},
  year={2024}
}

@inproceedings{chen2024monogaussianavatar,
  title={Monogaussianavatar: Monocular gaussian point-based head avatar},
  author={Chen, Yufan and Wang, Lizhen and Li, Qijing and Xiao, Hongjiang and Zhang, Shengping and Yao, Hongxun and Liu, Yebin},
  booktitle={ACM SIGGRAPH 2024 conference papers},
  pages={1--9},
  year={2024}
}

@article{wang2025gaussianhead,
  title={Gaussianhead: High-fidelity head avatars with learnable gaussian derivation},
  author={Wang, Jie and Xie, Jiu-Cheng and Li, Xianyan and Xu, Feng and Pun, Chi-Man and Gao, Hao},
  journal={IEEE Transactions on Visualization and Computer Graphics},
  year={2025},
  publisher={IEEE}
}

@inproceedings{huang2023simhmr,
  title={Simhmr: A simple query-based framework for parameterized human mesh reconstruction},
  author={Huang, Zihao and Shi, Min and Liu, Chengxin and Xian, Ke and Cao, Zhiguo},
  booktitle={Proceedings of the 31st ACM International Conference on Multimedia},
  pages={6918--6927},
  year={2023}
}

@inproceedings{wu2025cat4d,
  title={Cat4d: Create anything in 4d with multi-view video diffusion models},
  author={Wu, Rundi and Gao, Ruiqi and Poole, Ben and Trevithick, Alex and Zheng, Changxi and Barron, Jonathan T and Holynski, Aleksander},
  booktitle={Proceedings of the IEEE/CVF Conference on Computer Vision and Pattern Recognition},
  pages={26057--26068},
  year={2025}
}

@inproceedings{taubner2025cap4d,
  title={Cap4d: Creating animatable 4d portrait avatars with morphable multi-view diffusion models},
  author={Taubner, Felix and Zhang, Ruihang and Tuli, Mathieu and Lindell, David B},
  booktitle={2025 IEEE/CVF Conference on Computer Vision and Pattern Recognition (CVPR)},
  pages={5318--5330},
  year={2025},
  organization={IEEE Computer Society}
}

@article{kerbl20233d,
  title={3d gaussian splatting for real-time radiance field rendering.},
  author={Kerbl, Bernhard and Kopanas, Georgios and Leimk{\"u}hler, Thomas and Drettakis, George and others},
  journal={ACM Trans. Graph.},
  volume={42},
  number={4},
  pages={139--1},
  year={2023}
}

@inproceedings{deng2024portrait4d,
  title={Portrait4d: Learning one-shot 4d head avatar synthesis using synthetic data},
  author={Deng, Yu and Wang, Duomin and Ren, Xiaohang and Chen, Xingyu and Wang, Baoyuan},
  booktitle={Proceedings of the IEEE/CVF Conference on Computer Vision and Pattern Recognition},
  pages={7119--7130},
  year={2024}
}

@inproceedings{deng2024portrait4dv2,
  title={Portrait4d-v2: Pseudo multi-view data creates better 4d head synthesizer},
  author={Deng, Yu and Wang, Duomin and Wang, Baoyuan},
  booktitle={European Conference on Computer Vision},
  pages={316--333},
  year={2024},
  organization={Springer}
}

@article{ye2024real3d,
  title={Real3d-portrait: One-shot realistic 3d talking portrait synthesis},
  author={Ye, Zhenhui and Zhong, Tianyun and Ren, Yi and Yang, Jiaqi and Li, Weichuang and Huang, Jiawei and Jiang, Ziyue and He, Jinzheng and Huang, Rongjie and Liu, Jinglin and others},
  journal={arXiv preprint arXiv:2401.08503},
  year={2024}
}

@article{chu2024gpavatar,
  title={GPAvatar: Generalizable and precise head avatar from image (s)},
  author={Chu, Xuangeng and Li, Yu and Zeng, Ailing and Yang, Tianyu and Lin, Lijian and Liu, Yunfei and Harada, Tatsuya},
  journal={arXiv preprint arXiv:2401.10215},
  year={2024}
}

@inproceedings{li2023one,
  title={One-shot high-fidelity talking-head synthesis with deformable neural radiance field},
  author={Li, Weichuang and Zhang, Longhao and Wang, Dong and Zhao, Bin and Wang, Zhigang and Chen, Mulin and Zhang, Bang and Wang, Zhongjian and Bo, Liefeng and Li, Xuelong},
  booktitle={Proceedings of the IEEE/CVF Conference on Computer Vision and Pattern Recognition},
  pages={17969--17978},
  year={2023}
}

@article{li2023generalizable,
  title={Generalizable one-shot 3d neural head avatar},
  author={Li, Xueting and De Mello, Shalini and Liu, Sifei and Nagano, Koki and Iqbal, Umar and Kautz, Jan},
  journal={Advances in Neural Information Processing Systems},
  volume={36},
  pages={47239--47250},
  year={2023}
}

@inproceedings{ma2024cvthead,
  title={Cvthead: One-shot controllable head avatar with vertex-feature transformer},
  author={Ma, Haoyu and Zhang, Tong and Sun, Shanlin and Yan, Xiangyi and Han, Kun and Xie, Xiaohui},
  booktitle={Proceedings of the IEEE/CVF Winter Conference on Applications of Computer Vision},
  pages={6131--6141},
  year={2024}
}

@inproceedings{feng2025gpavatar,
  title={Gpavatar: High-fidelity head avatars by learning efficient gaussian projections},
  author={Feng, Wei-Qi and Han, Dong and Zhou, Ze-Kang and Li, Shunkai and Liu, Xiaoqiang and Wan, Pengfei and Zhang, Di and Wang, Miao},
  booktitle={Proceedings of the Computer Vision and Pattern Recognition Conference},
  pages={250--259},
  year={2025}
}

@article{liang2025fastavatar,
  title={FastAvatar: Instant 3D Gaussian Splatting for Faces from Single Unconstrained Poses},
  author={Liang, Hao and Ge, Zhixuan and Majee, Soumendu and Tiwari, Ashish and Godaliyadda, GM and Veeraraghavan, Ashok and Balakrishnan, Guha},
  journal={arXiv preprint arXiv:2508.18389},
  year={2025}
}

@inproceedings{chu2025artalk,
  title={Artalk: Speech-driven 3d head animation via autoregressive model},
  author={Chu, Xuangeng and Goswami, Nabarun and Cui, Ziteng and Wang, Hanqin and Harada, Tatsuya},
  booktitle={Proceedings of the SIGGRAPH Asia 2025 Conference Papers},
  pages={1--9},
  year={2025}
}

@article{simeoni2025dinov3,
  title={Dinov3},
  author={Sim{\'e}oni, Oriane and Vo, Huy V and Seitzer, Maximilian and Baldassarre, Federico and Oquab, Maxime and Jose, Cijo and Khalidov, Vasil and Szafraniec, Marc and Yi, Seungeun and Ramamonjisoa, Micha{\"e}l and others},
  journal={arXiv preprint arXiv:2508.10104},
  year={2025}
}

@inproceedings{xie2022vfhq,
  title={Vfhq: A high-quality dataset and benchmark for video face super-resolution},
  author={Xie, Liangbin and Wang, Xintao and Zhang, Honglun and Dong, Chao and Shan, Ying},
  booktitle={Proceedings of the IEEE/CVF Conference on Computer Vision and Pattern Recognition},
  pages={657--666},
  year={2022}
}

@article{kirschstein2023nersemble,
  title={Nersemble: Multi-view radiance field reconstruction of human heads},
  author={Kirschstein, Tobias and Qian, Shenhan and Giebenhain, Simon and Walter, Tim and Nie{\ss}ner, Matthias},
  journal={ACM Transactions on Graphics (TOG)},
  volume={42},
  number={4},
  pages={1--14},
  year={2023},
  publisher={ACM New York, NY, USA}
}

@article{wang2004image,
  title={Image quality assessment: from error visibility to structural similarity},
  author={Wang, Zhou and Bovik, Alan C and Sheikh, Hamid R and Simoncelli, Eero P},
  journal={IEEE transactions on image processing},
  volume={13},
  number={4},
  pages={600--612},
  year={2004},
  publisher={IEEE}
}

@inproceedings{zhang2018unreasonable,
  title={The unreasonable effectiveness of deep features as a perceptual metric},
  author={Zhang, Richard and Isola, Phillip and Efros, Alexei A and Shechtman, Eli and Wang, Oliver},
  booktitle={Proceedings of the IEEE conference on computer vision and pattern recognition},
  pages={586--595},
  year={2018}
}

@inproceedings{deng2019arcface,
  title={Arcface: Additive angular margin loss for deep face recognition},
  author={Deng, Jiankang and Guo, Jia and Xue, Niannan and Zafeiriou, Stefanos},
  booktitle={Proceedings of the IEEE/CVF conference on computer vision and pattern recognition},
  pages={4690--4699},
  year={2019}
}

@article{li2017learning,
  title={Learning a model of facial shape and expression from 4D scans.},
  author={Li, Tianye and Bolkart, Timo and Black, Michael J and Li, Hao and Romero, Javier},
  journal={ACM Trans. Graph.},
  volume={36},
  number={6},
  pages={194--1},
  year={2017}
}

@article{wu2025fastavatar,
  title={FastAvatar: Towards Unified and Fast 3D Avatar Reconstruction with Large Gaussian Reconstruction Transformers},
  author={Wu, Yue and Chen, Xuanhong and Wu, Yufan and Li, Wen and Lu, Yuxi and Feng, Kairui},
  journal={arXiv preprint arXiv:2508.19754},
  year={2025}
}

@inproceedings{ji2026fastgha,
  title={FastGHA: Generalized Few-Shot 3D Gaussian Head Avatars with Real-Time Animation},
  author={Ji, Xinya and Weiss, Sebastian and Kansy, Manuel and Naruniec, Jacek and Cao, Xun and Solenthaler, Barbara and Bradley, Derek},
  booktitle={The Fourteenth International Conference on Learning Representations},
  year={2026}
}

@inproceedings{xu2024gaussian,
  title={Gaussian head avatar: Ultra high-fidelity head avatar via dynamic gaussians},
  author={Xu, Yuelang and Chen, Benwang and Li, Zhe and Zhang, Hongwen and Wang, Lizhen and Zheng, Zerong and Liu, Yebin},
  booktitle={Proceedings of the IEEE/CVF conference on computer vision and pattern recognition},
  pages={1931--1941},
  year={2024}
}

@article{martinez2024codec,
  title={Codec avatar studio: Paired human captures for complete, driveable, and generalizable avatars},
  author={Martinez, Julieta and Kim, Emily and Romero, Javier and Bagautdinov, Timur and Saito, Shunsuke and Yu, Shoou-I and Anderson, Stuart and Zollh{\"o}fer, Michael and Wang, Te-Li and Bai, Shaojie and others},
  journal={Advances in Neural Information Processing Systems},
  volume={37},
  pages={83008--83023},
  year={2024}
}

\clearpage
\appendix
\setcounter{section}{0}
\setcounter{subsection}{0}
\setcounter{figure}{0}
\setcounter{table}{0}
\renewcommand{\thesection}{\Alph{section}}
\renewcommand{\thesubsection}{\thesection.\arabic{subsection}}
\renewcommand{\thefigure}{S\arabic{figure}}
\renewcommand{\thetable}{S\arabic{table}}

\section*{Supplementary Material}

This supplementary document provides additional implementation details, experimental results, qualitative analyses, and discussions that complement the main paper. 
Specifically, Section~\ref{sec:additional_details} presents additional implementation details, 
Section~\ref{sec:additional_quantitative} reports additional quantitative results, including runtime analysis and ablation studies,
Section~\ref{sec:more_Qualitative} provides further qualitative comparisons, 
Section~\ref{sec:application} presents several related applications of our framework,
Section~\ref{sec:robu} analyzes the robustness of the proposed method under challenging input conditions, 
Section~\ref{sec:ethical} discusses ethical considerations, and 
Section~\ref{sec:limitation} summarizes the limitations of our approach and outlines potential directions for future work. 
In addition, we provide a supplementary video to better illustrate the reconstruction and animation results.

\section{Additional Implementation Details}
\label{sec:additional_details}

\subsection{Training and Testing Details}

Our model is trained on 8 NVIDIA Tesla A800 GPUs using a two-stage training strategy. We first train all modules except the motion-aware refinement module, and then freeze the remaining modules to optimize only the motion-aware refinement module. The overall training process takes approximately one week. We evaluate the model on a single RTX 4090D GPU with 24GB memory.
For evaluation, we use the official VFHQ~\cite{xie2022vfhq} test split, which contains 50 identities, as well as four identity sequences from the NeRSemble~\cite{kirschstein2023nersemble} dataset.
Following the same preprocessing pipeline as in training, we first detect facial regions and crop the images to a resolution of $512 \times 512$ pixels.
For each frame, camera pose estimation and FLAME~\cite{li2017learning} parameter tracking are performed to obtain head pose and facial expression parameters.
In addition, we perform background removal to separate the subject from the background. The background region is filled with a white color to emphasize the foreground subject and reduce background interference during training and evaluation.

\begin{table}[t]
\caption{\textbf{Main hyper-parameters of the FFAvatar architecture.} 
The alternating-attention aggregator extracts and aggregates multi-frame visual tokens, while the cross-modal alignment transformer and the motion refinement transformer decode static identity features and dynamic motion-dependent residual features, respectively.}
\label{tab:ffavatar_hparams}
\centering
\small
\renewcommand{\arraystretch}{1.15}
\setlength{\tabcolsep}{4pt}
\resizebox{\linewidth}{!}{
\begin{tabular}{l|l|l}
\hline
\textbf{Module} & \textbf{Hyper-parameter} & \textbf{Setting} \\
\hline

\multirow{13}{*}{Alternating Attention Aggregator}
& Input resolution & $512 \times 512$ \\
& Backbone & DINOv3 ViT-L/16 (Freeze)\\
& Patch size & $16$ \\
& Spatial token resolution & $32 \times 32$ \\
& Token dimension & $1024$ \\
& Number of attention blocks & $24$ \\
& Number of attention heads & $16$ \\
& MLP ratio & $4.0$ \\
& Attention order & [frame, global] \\
& Attention switch interval & $1$ block (\texttt{aa\_block\_size}=1) \\
& QK normalization & enabled \\
& Register tokens & $4$ \\
& Positional encoding & 2D rotary embedding, frequency $=100$ \\
\hline

\multirow{7}{*}{Sparse-to-Dense Cross-modal Alignment}
& Block type & transfromer \\
& Number of layers & $8$ (sparse) + $2$ (dense) \\
& UV Resolution & $128$ \\
& Number of attention heads & $16$ \\
& Hidden dimension & $1024$ \\
& Conditioning dimension & $1024$ \\
& Gradient checkpointing & enabled \\
\hline

\multirow{6}{*}{Motion-Aware Refinement}
& Block type & transfromer \\
& Number of layers & $4$ \\
& Number of attention heads & $8$ \\
& Hidden dimension & $1024$ \\
& Conditioning dimension & $1024$ \\
& Gradient checkpointing & enabled \\
\hline
\multirow{4}{*}{GS Decoder}
& Input dimension & $1024$ \\
& Output attributes & $\Delta\mathbf{x}$ (3), $\mathbf{s}$ (3), $\mathbf{q}$ (4), $\alpha$ (1), RGB (3) \\
& Scaling clipping & $0.2$ \\
& Position offset range & $[-0.1, 0.1]$ \\
\hline
\end{tabular}}

\end{table}

\subsection{Network Architecture Details}

Our framework consists of three components: an alternating-attention visual encoder, a sparse-to-dense cross-modal alignment module, and a motion refinement module.

\noindent\textbf{Alternating-attention visual encoder.}
We use a frozen DINOv3 ViT-L/16~\cite{simeoni2025dinov3} backbone as the image feature extractor. Given an input image of size $3 \times 512 \times 512$, the backbone produces a $32 \times 32$ patch grid, i.e., 1024 visual tokens. Multi-scale features from four intermediate layers $\{4,11,17,23\}$ are fused by a DPT-style head, yielding 1024-dimensional dense features. On top of these features, we apply an alternating-attention~\cite{wang2025vggt} aggregator with token dimension 1024, depth 24, 16 attention heads, and MLP ratio 4.0. The attention order is set to \texttt{[frame, global]} with \texttt{aa\_block\_size}=1. We additionally use 4 register tokens, QK normalization, and 2D rotary positional encoding with frequency 100. Expression and pose parameters are projected to 1024-dimensional tokens and concatenated with visual tokens before aggregation.

\noindent\textbf{Sparse-to-dense cross-modal alignment module.}
The aligned 3D appearance is predicted by a conditional Transformer defined on FLAME-based~\cite{li2017learning} point queries. 
The architecture consists of 10 Transformer layers, including 8 sparse layers operating on FLAME vertex queries and 2 dense layers operating on UV-space~\cite{jiang2025uv,xiang2024flashavatar} samples. 
Each layer uses 16 attention heads with a hidden dimension of 1024. 
The network takes point embeddings as queries and aggregated image tokens as conditioning features, and predicts point-aligned latent features for canonical Gaussian avatar construction. 

\noindent\textbf{Motion-Aware refinement module.}
To model personalized non-rigid motion beyond the FLAME~\cite{li2017learning} prior, we use a lightweight residual transformer. This module adopts the same hidden dimension 1024, but uses 4 layers and 8 attention heads. It is conditioned on pose and expression tokens and predicts motion-dependent residual features, which are converted into residual Gaussian attributes and fused with the canonical Gaussian~\cite{kerbl20233d} representation for final rendering.

\noindent\textbf{GS decoder.}
The GS decoder converts per-point latent features into the final Gaussian primitive attributes that constitute the drivable 3D head avatar.
It takes as input the 1024-dimensional features output by the motion-aware refinement module and decodes each Gaussian attribute through an independent linear projection layer $\mathrm{Linear}(1024, d_{\text{out}})$.
The decoded attributes include a per-primitive position offset $\Delta\mathbf{x} \in \mathbb{R}^{3}$ from the FLAME vertices, anisotropic scale $\mathbf{s} \in \mathbb{R}^{3}$, rotation quaternion $\mathbf{q} \in \mathbb{R}^{4}$, opacity $\alpha \in \mathbb{R}$, and RGB color $\mathbf{c} \in \mathbb{R}^{3}$.
The position offset is bounded to $[-0.1, 0.1]$ via a scaled sigmoid, scale is activated by a truncated exponential with a clipping upper bound of $0.2$, opacity uses a sigmoid with initial bias set to $\sigma^{-1}(0.1)$, and the rotation quaternion is $\ell_2$-normalized to ensure validity.

\subsection{Detailed Loss Formulation}
\label{sec:supp_loss}
The model is trained using photometric supervision and geometric regularization. 
The overall objective is
\[
\mathcal{L}
=
\mathcal{L}_{\mathrm{rgb}}
+
\lambda_{\mathrm{mask}} \mathcal{L}_{\mathrm{mask}}
+
\lambda_{\mathrm{off}} \mathcal{L}_{\mathrm{off}} .
\]
\noindent\textbf{RGB Loss.}
We combine pixel reconstruction and perceptual similarity:
\[
\mathcal{L}_{\mathrm{rgb}}
=
\left\|
\mathbf{I}^{\mathrm{pred}}
-
\mathbf{I}^{\mathrm{gt}}
\right\|_1
+
\lambda_{\mathrm{lpips}}
\,
\mathrm{LPIPS}
\left(
\mathbf{I}^{\mathrm{pred}},
\mathbf{I}^{\mathrm{gt}}
\right).
\]
\noindent\textbf{Mask Loss.}
Foreground alignment is supervised using
\[
\mathcal{L}_{\mathrm{mask}}
=
\left\|
\mathbf{S}^{\mathrm{pred}}
-
\mathbf{S}^{\mathrm{gt}}
\right\|_1 .
\]

The mask weight gradually increases during training:

\[
\lambda_{\mathrm{mask}}(t)
=
\lambda_{\mathrm{start}}
+
(\lambda_{\mathrm{end}}-\lambda_{\mathrm{start}})
\frac{1-e^{-\alpha t}}{1-e^{-\alpha}},
\]

where $t=\min(\text{step}/T,1)$. We use $\lambda_{\mathrm{start}}=0$, $\lambda_{\mathrm{end}}=1.0$, $\alpha=0.5$, and $T=10{,}000$.

\noindent\textbf{Offset Regularization.}
To stabilize Gaussian deformation, we regularize the predicted center offsets:

\[
\mathcal{L}_{\mathrm{off}}
=
\frac{1}{|\mathcal{P}|}
\sum_{\mathbf{p}\in\mathcal{P}}
\left\|
\Delta \mathbf{p}
\right\|_2 .
\]

\noindent\textbf{Loss Weights.}
We set $\lambda_{\mathrm{lpips}}=1.0$ and $\lambda_{\mathrm{off}}=0.1$.

\begin{table}[t]
\centering
\setlength{\tabcolsep}{2pt}

\begin{minipage}{0.65\columnwidth}
\centering
\caption{
$^{\#}$ tested on RTX 4090D;
$^{*}$ tested on A800.
}
\label{tab:fps}
\scriptsize
\resizebox{\linewidth}{!}{
\begin{tabular}{c|cccccc}
\toprule
\multirow{2}{*}{Method} &
\multicolumn{6}{c}{\textcolor{red}{Creation Time$\downarrow$ (s) }/ \textcolor{blue}{FPS$\uparrow$ (img/s)}} \\
\cmidrule(lr){2-7}
& 1$^{\#}$ & 4$^{\#}$ & 8$^{\#}$ & 16$^{\#}$ & 32$^{\#}$ & 108$^{*}$ \\
\midrule
    FastAvatar & \textcolor{red}{2.6}/\textcolor{blue}{339} & \textcolor{red}{6.6}/\textcolor{blue}{167} & \textcolor{red}{12.2}/\textcolor{blue}{97} & \textcolor{red}{25.1}/\textcolor{blue}{51} &  - & - \\
    Ours & \textcolor{red}{1.3}/\textcolor{blue}{31} & \textcolor{red}{1.7}/\textcolor{blue}{31} & \textcolor{red}{2.1}/\textcolor{blue}{31} & \textcolor{red}{4.3}/\textcolor{blue}{31} & \textcolor{red}{11.6}/\textcolor{blue}{31} & \textcolor{red}{122.3}/\textcolor{blue}{23} \\
    Ours(fast) & \textcolor{red}{1.3}/\textcolor{blue}{468} & \textcolor{red}{1.7}/\textcolor{blue}{468} & \textcolor{red}{2.1}/\textcolor{blue}{468} & \textcolor{red}{4.3}/\textcolor{blue}{468} & \textcolor{red}{11.6}/\textcolor{blue}{468} & \textcolor{red}{122.3}/\textcolor{blue}{229} \\

\bottomrule
\end{tabular}
}
\end{minipage}
\hfill
\begin{minipage}{0.34\columnwidth}
\centering
\caption{Ablation studies.}
\label{tab:ablation_more}
\scriptsize
\resizebox{\linewidth}{!}{
\begin{tabular}{l|ccc}
\toprule
Method & PSNR$\uparrow$ & SSIM$\uparrow$ & LPIPS$\downarrow$ \\
\midrule

UV-only & 23.05 & 0.861 & 0.092 \\
w/o Cam. & 23.33 & 0.864 & 0.080 \\
Ours & \textbf{23.35} & \textbf{0.864} & \textbf{0.081} \\

\bottomrule
\end{tabular}
}
\end{minipage}

\end{table}

\section{Additional Quantitative Results}
\label{sec:additional_quantitative}

\subsection{Runtime and Scalability Analysis}

Table~\ref{tab:fps} reports the avatar creation time and animation speed under different numbers of reference images.
Compared with FastAvatar~\cite{wu2025fastavatar}, which reconstructs and fuses view-wise Gaussian representations, FFAvatar directly predicts a unified canonical Gaussian field.
As a result, the number of Gaussians, rendering cost, and animation speed are independent of the number of input images.
Although the avatar creation time and GPU memory consumption increase with more reference images, our method can scale to more than one hundred input frames on larger-memory GPUs, such as NVIDIA A800.
In practice, avatar creation is performed only once, while the reconstructed avatar can be animated repeatedly.
Therefore, we prioritize real-time animation speed after avatar creation.

Our standard setting achieves real-time animation on a single RTX 4090D GPU, while the fast setting further improves the rendering speed. These results demonstrate the scalability of the proposed canonical representation and its potential for real-time avatar animation applications.

\subsection{Additional Ablation Studies}

Table~\ref{tab:ablation_more} provides additional ablation results on the representation design and camera conditioning.
Compared with UV-only Gaussians, the full canonical Gaussian representation achieves better performance by combining the expanded UV-space features with the original FLAME point features, which helps preserve point-level geometric cues during sparse-to-dense feature expansion and provides a denser 3D Gaussian representation.
We also evaluate the effect of camera conditioning.
Removing camera parameters leads to comparable image-level metrics, but camera conditioning provides explicit pose information and helps reduce the ambiguity between head pose and facial motion.
Therefore, we keep camera parameters as auxiliary conditioning in the full model.
Our current framework does not explicitly model view-dependent appearance effects, which remains a limitation and an interesting direction for future work.

\subsection{Comparison with Real-Time Feed-Forward Frameworks}

FFAvatar is designed to balance high-fidelity reconstruction and efficient animation.
Unlike methods whose rendering or animation cost depends on view-wise or input-dependent representations, such as Avat3r~\cite{kirschstein2025avat3r}, FastAvatar~\cite{wu2025fastavatar}, and FastGHA~\cite{ji2026fastgha}, our method adopts a unified canonical Gaussian representation.
This design makes the animation speed independent of the number of reference images, as shown in Table~\ref{tab:fps}.
In addition, the fast setting achieves substantially higher animation speed while using the same reconstructed avatar, further demonstrating the potential of FFAvatar for real-time AR/VR applications.
Together with the quantitative and qualitative comparisons in the main paper, these results show that FFAvatar provides a favorable trade-off between reconstruction quality, input scalability, and animation efficiency.

\section{Additional Qualitative Results}
\label{sec:more_Qualitative}
In this section, we provide additional qualitative comparisons to further analyze the behavior of FFAvatar under different animation scenarios. The results highlight the advantages of our approach in terms of identity preservation, expression controllability, and robustness under challenging conditions.

\begin{figure}[!t]
\centering
\includegraphics[width=1.0\linewidth]{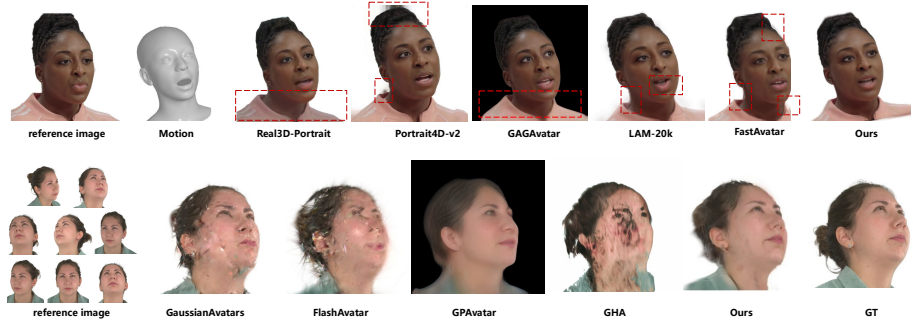}
\caption{
\textbf{Comparison of novel expression and viewpoint synthesis.}
We compare our method with Real3D-Portrait~\cite{ye2024real3d}, Portrait4D-v2~\cite{deng2024portrait4dv2}, GAGAvatar~\cite{chu2024generalizable}, LAM~\cite{he2025lam}, and FastAvatar~\cite{wu2025fastavatar} under the single-reference setting. 
Under the few-reference setting, we further include GPAvatar~\cite{chu2024gpavatar} and the optimization-based methods GaussianAvatars~\cite{qian2024gaussianavatars}, FlashAvatar~\cite{xiang2024flashavatar}, and GHA~\cite{xu2024gaussian}. 
Our method preserves finer facial details while maintaining more consistent geometry and appearance across novel expressions and viewpoints, particularly under large viewpoint changes.
\label{fig:supp_exp_view}
}
\end{figure}

\begin{figure}[!t]
\centering
\includegraphics[width=1.0\linewidth]{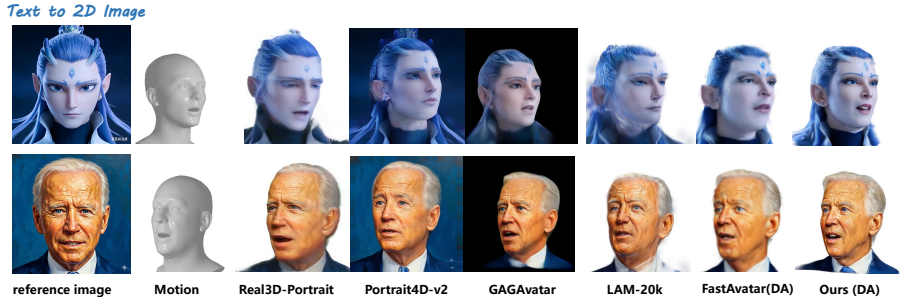}
\caption{\label{fig:supp_os}
\textbf{Open-scenario avatar reconstruction.}
Qualitative results of avatar reconstruction from stylized portraits generated by text prompts. With diffusion-based augmentation (DA) for enhanced input features, FFAvatar remains robust to out-of-domain inputs and preserves coherent identity appearance under both novel expression and novel view synthesis.}
\end{figure}

\begin{figure}[!t]
\centering
\includegraphics[width=1.0\linewidth]{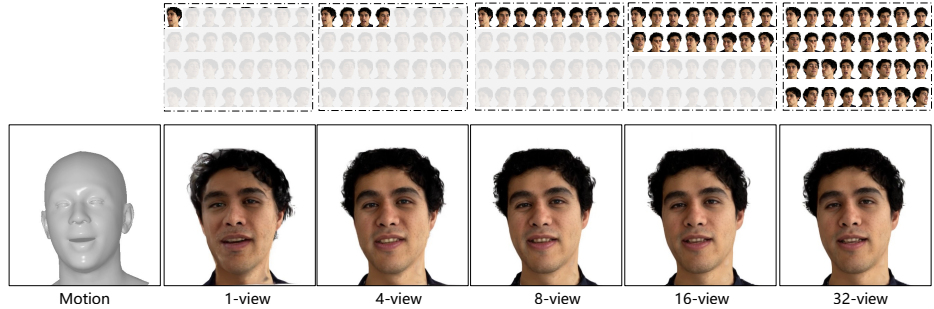}
\caption{\label{fig:supp_1to32_results}
\textbf{Qualitative comparison with different numbers of reference views.} The dashed boxes visualize the selected reference portraits for each setting, and the first column shows the driving motion. We compare avatar reconstruction results using 1, 4, 8, 16, and 32 reference views, following the multi-view settings evaluated in Table~3 of the main paper.}
\end{figure}

\textbf{Novel Expression and View Synthesis.}
Fig.~\ref{fig:supp_exp_view} presents qualitative comparisons for novel expression and viewpoint synthesis under both single-reference and few-reference settings.
For novel expression synthesis, we compare with Real3D-Portrait~\cite{ye2024real3d}, Portrait4D-v2~\cite{deng2024portrait4dv2}, GAGAvatar~\cite{chu2024generalizable}, LAM-20k~\cite{he2025lam}, and FastAvatar~\cite{wu2025fastavatar}.
LAM-20k produces inaccurate mouth motions, while GAGAvatar exhibits artifacts below the head and around the neck.
Portrait4D-v2 and FastAvatar generate noticeable artifacts around the head region, whereas Real3D-Portrait produces over-smoothed results, particularly below the head.
Our method reconstructs finer head details while preserving more consistent identity and appearance.

For novel viewpoint synthesis, we further compare with GPAvatar~\cite{chu2024gpavatar} and the optimization-based methods GaussianAvatars~\cite{qian2024gaussianavatars}, FlashAvatar~\cite{xiang2024flashavatar}, and GHA~\cite{xu2024gaussian}.
Compared with previous approaches, our method maintains more consistent geometry and appearance under large viewpoint changes.

\textbf{Open-Scenario Avatar Reconstruction.}
Fig.~\ref{fig:supp_os} shows results under open-scenario settings, where the input portraits are generated from text prompts, such as ``a portrait of Joe Biden'' or ``an anime-style avatar of Ao Yin from the movie Ne Zha''. 
These stylized images are then used as inputs to different avatar reconstruction methods. 
By leveraging diffusion-based augmentation~\cite{taubner2025cap4d} to enhance input features, FFAvatar exhibits stronger robustness to out-of-domain portraits and preserves coherent identity appearance under both novel expression and novel view synthesis.

\textbf{Effect of the Number of Reference Views.} To complement the multi-view quantitative results reported in Table~3 of the main paper, Fig.~\ref{fig:supp_1to32_results} provides qualitative comparisons under different numbers of reference views. We reconstruct the avatar using 1, 4, 8, 16, and 32 input views while applying the same driving motion. With only a single reference view, FFAvatar already produces a plausible reconstruction. As the number of reference views increases, the reconstructed avatar exhibits more complete and consistent identity appearance, especially in facial shape, hair contour, and side-view regions. The results under 16 and 32 views are visually similar, suggesting that FFAvatar can effectively exploit multi-view inputs while maintaining stable reconstruction quality.

\begin{figure}[t]
\centering
\includegraphics[width=1.0\linewidth]{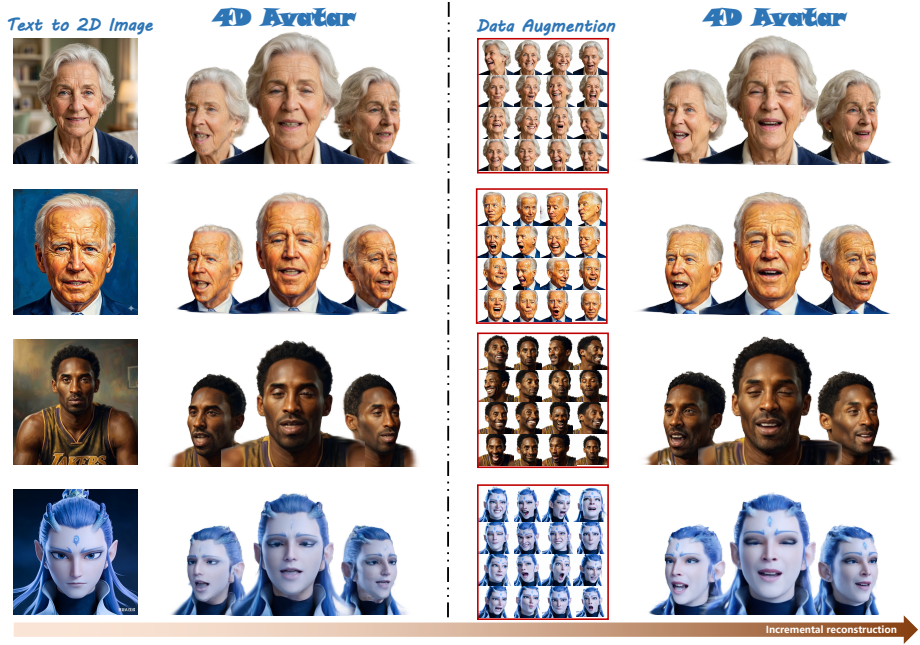}
\caption{\textbf{Text-to-4D avatar generation.}
We first generate reference images from text prompts using a text-to-image model, and then reconstruct a dynamic 4D head avatar using FFAvatar. 
Benefiting from diffusion-based data augmentation and the incremental reconstruction strategy, our method produces avatars with consistent appearance across viewpoints and realistic facial motion.
\label{fig:text_to_4d}}
\end{figure}

\section{Related Applications}
\label{sec:application}

Beyond high-quality avatar reconstruction, the proposed framework also enables several practical applications due to its flexible feed-forward design and disentangled representation of identity and motion.

\textbf{(1) Text-to-4D Avatar.}
As shown in Fig.~\ref{fig:text_to_4d}, our framework can be used to generate a 4D head avatar directly from text prompts. 
Specifically, we first generate reference images using a text-to-image model, and then feed these images into FFAvatar to reconstruct a dynamic 4D avatar. 
Benefiting from diffusion-based~\cite{taubner2025cap4d} data augmentation and the incremental reconstruction strategy, our method produces avatars with improved view consistency and more realistic motion when synthesizing novel viewpoints and expressions.

\begin{figure}[!t]
\centering
\includegraphics[width=1.0\linewidth]{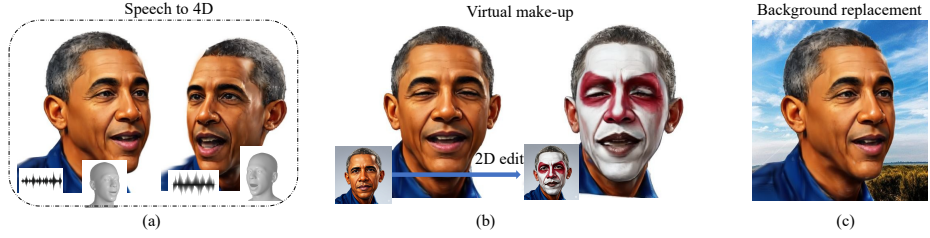}
\caption{\textbf{Related applications of FFAvatar.}
(a) Speech-driven talking head generation by combining FFAvatar with audio-to-3DMM methods~\cite{chu2025artalk}. 
(b) Avatar appearance editing through modifications of the input reference images. 
(c) Background replacement enabled by the explicit 3D Gaussian representation, allowing the avatar to be rendered under arbitrary scenes.
\label{fig:supp_extent}}
\end{figure}

\textbf{(2) Speech-driven Talking Head.}
Speech-driven talking head synthesis is an important application in digital humans and virtual agents. 
As illustrated in Fig.~\ref{fig:supp_extent}(a), FFAvatar can be combined with existing audio-to-3DMM approaches~\cite{chu2025artalk} to enable speech-driven animation. 
Given audio signals, the driving method predicts FLAME expression and pose parameters, which are then used to animate the reconstructed avatar. 
This design allows our method to generate expressive and identity-consistent talking head animations while preserving high rendering quality.

\textbf{(3) Image Editing and Background Replacement.}
Our framework also supports flexible avatar editing through modifications of the reference images. 
As shown in Fig.~\ref{fig:supp_extent}(b), editing the 2D input images (e.g., changing color or style) can indirectly control the appearance of the reconstructed 4D avatar, enabling style transfer and appearance editing. 
Furthermore, unlike some previous methods that rely on post-processing networks to refine rendered results~\cite{chu2024generalizable,chu2024gpavatar}, our approach is based on an explicit 3D representation. 
This allows straightforward background replacement by manipulating Gaussian opacity values, as illustrated in Fig.~\ref{fig:supp_extent}(c), enabling the avatar to be rendered under arbitrary backgrounds.

\begin{figure}[!t]
\centering
\includegraphics[width=1.0\linewidth]{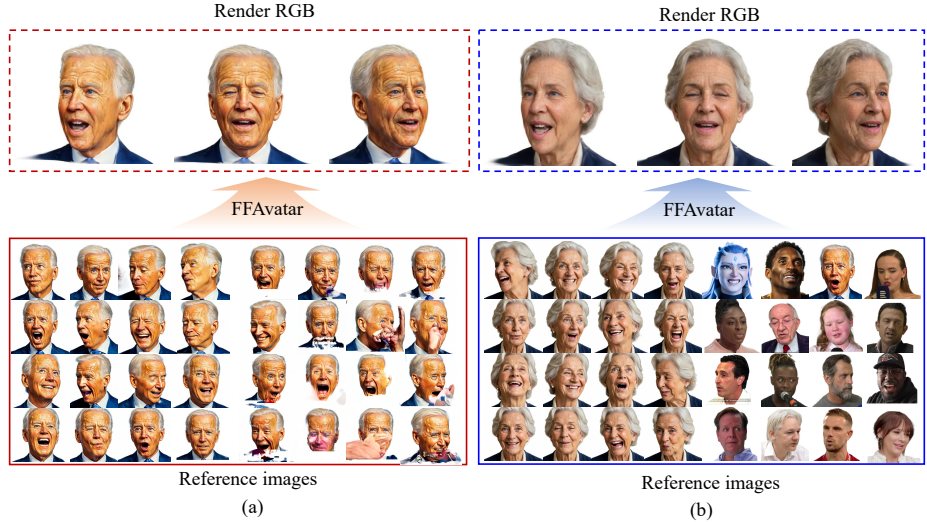}
\caption{
\textbf{Robustness to noisy and irrelevant inputs.}
(a) Reconstruction results using 32 reference views, where 16 views contain severe degradations such as occlusions, missing facial regions, geometric distortions, and color shifts. 
(b) Results under a more extreme setting where half of the inputs correspond to completely unrelated identities. 
FFAvatar remains stable and preserves consistent identity despite the presence of corrupted or distracting inputs.
\label{fig:supp_robu}
}
\end{figure}
\section{Robustness Analysis}
\label{sec:robu}

We further evaluate the robustness of FFAvatar under challenging few-shot input conditions. 
As shown in Fig.~\ref{fig:supp_robu}(a), we use 32 reference views as the model input, where half of them are intentionally degraded low-quality samples. 
These degraded inputs include severe artifacts such as missing facial regions, occlusions, geometric distortions, and color shifts. 
Despite the presence of these corrupted views, our method is still able to reconstruct a stable and high-quality 4D avatar while preserving consistent identity appearance.
To further stress-test the robustness of the model, we construct a more extreme scenario. 
As illustrated in Fig.~\ref{fig:supp_robu}(b), we again provide 32 reference views as input, but intentionally replace half of them with images that contain completely unrelated identity information. 
Surprisingly, the reconstructed avatar remains largely unaffected by these distracting inputs, and the rendered results still maintain the correct identity and stable appearance.
These qualitative results demonstrate that FFAvatar is highly robust to noisy and misleading inputs. 
The model can effectively filter out corrupted or irrelevant information and focus on consistent identity cues across the reference images, making it suitable for real-world scenarios where input data may be noisy, incomplete, or partially mismatched.



\section{Ethical Considerations}
\label{sec:ethical}
While our method enables efficient avatar creation, it may also raise potential concerns related to misuse of digital identities. 
To mitigate such risks, we emphasize that the technology should be used in compliance with applicable laws and ethical guidelines.
Future work may explore watermarking and identity verification mechanisms to improve responsible deployment.

\section{Limitations and Future Work}
\label{sec:limitation}

\begin{figure}[!t]
\centering
\includegraphics[width=1.0\linewidth]{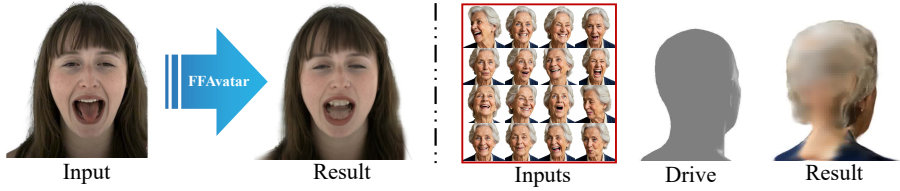}
\caption{
\textbf{Failure cases.}
The tongue is not explicitly modeled, and unseen regions from the input views may contain limited details under extreme novel-view rendering.
}
\label{fig:limitation}
\end{figure}
Although FFAvatar demonstrates strong performance across multiple benchmarks, several limitations remain.
First, our method can reconstruct challenging attributes such as glasses and complex hairstyles in many cases, but it may still struggle with highly dynamic or weakly observed structures.
As shown in Fig.~\ref{fig:limitation}, the current representation does not explicitly model the tongue, which may lead to inaccurate mouth interiors for large mouth-opening expressions.
In addition, regions that are rarely or never observed in the reference images, such as the back of the head under front-view inputs, may contain fewer fine-grained details when rendered from extreme side or rear viewpoints.
Second, although FFAvatar supports reconstruction from one or more reference portrait images, single-view inputs remain more challenging than multi-view or multi-expression inputs.
When only one reference image is available, ambiguities in unseen geometry and appearance may cause view-dependent artifacts, especially under large pose changes or turntable-style rendering.
Using additional reference images with complementary viewpoints and expressions can alleviate this issue.
Third, our framework relies on the FLAME parametric model and tracking pipeline for motion control.
Therefore, animation accuracy is influenced by the quality of estimated FLAME parameters.
Inaccurate tracking may result in imperfect expression transfer or pose reconstruction, particularly for extreme expressions, rapid motion, and partially occluded faces.
Finally, our training data may inherit demographic biases from VFHQ~\cite{xie2022vfhq}.
Since VFHQ is collected mainly from in-the-wild interview videos, the distribution of identities, skin tones, ages, and facial appearances may be imbalanced.
This potential bias may affect reconstruction quality and robustness for underrepresented groups or uncommon appearance patterns.
Future work will explore more balanced training data, improved modeling of dynamic facial components, more robust handling of unseen regions, and extensions to dynamic hair and full-body avatar reconstruction.

%
%

\end{document}